%% file: neurips_2026.tex
\documentclass{article}

    \PassOptionsToPackage{numbers, compress}{natbib}

\usepackage[main, final]{neurips_2026}

\usepackage[utf8]{inputenc} 
\usepackage[T1]{fontenc}    

\usepackage{amsmath}
\usepackage{amssymb}
\usepackage{amsfonts}       
\usepackage{mathtools}
\usepackage{amsthm}

\usepackage{booktabs}       
\usepackage{nicefrac}       
\usepackage{microtype}      
\usepackage{xcolor}         
\usepackage{graphicx}
\usepackage{subcaption}
\usepackage{multirow}
\usepackage{makecell}
\usepackage{enumitem}
\usepackage{soul}
\usepackage[normalem]{ulem}
\usepackage{algorithm}
\usepackage{algpseudocode}
\usepackage{wrapfig}
\usepackage{paracol}
\usepackage[textsize=tiny]{todonotes}

\usepackage{url}            

\usepackage{hyperref}

\usepackage[capitalize,noabbrev]{cleveref}

\usepackage{xspace}

\newcommand{\OurMethod}{\texttt{CAPE}\xspace}
\DeclareMathOperator*{\argmin}{arg\,min}
\theoremstyle{plain}
\newtheorem{theorem}{Theorem}[section]

\theoremstyle{definition}

\theoremstyle{remark}

\allowdisplaybreaks[1]

\title{Contrastive Nonmyopic Objective Cost-Tradeoff Acquisition for Longitudinal Data}

\author{%
  Dzung Dinh\thanks{Equal contribution.} \\
  Dept. of Computer Science \\
  UNC Chapel Hill \\
  \texttt{ddinh@cs.unc.edu}
  \And
  Boqi Chen\footnotemark[1] \\
  Dept. of Computer Science \\
  UNC Chapel Hill \\
  \texttt{bqchen@cs.unc.edu}
  \And
  Yunni Qu \\
  Dept. of Computer Science \\
  UNC Chapel Hill \\
  \texttt{quyunni@cs.unc.edu}
  \AND
  Marc Niethammer \\
  Dept. of Computer Science and Engineering \\
  UC San Diego \\
  \texttt{mniethammer@ucsd.edu}
  \And
  Junier Oliva \\
  Dept. of Computer Science \\
  UNC Chapel Hill \\
  \texttt{joliva@cs.unc.edu}
}

\begin{document}

\maketitle

\begin{abstract}
In many critical domains, features are not freely available at inference time: each measurement may come with a cost of time, money, and risk. Longitudinal prediction further complicates this setting because both features and labels evolve over time, and measurements missed at earlier timepoints may become permanently unavailable. We propose \OurMethod (\textbf{C}ontrastive \textbf{A}cquisition-\textbf{P}lan \textbf{E}mbeddings), a partitioning acquisition policy for longitudinal active feature acquisition. \OurMethod first defines NOCT, an oracle-based objective that scores a set of future feature-time acquisitions by its expected predictive loss together with its acquisition cost. Based on NOCT, \OurMethod learns a contrastive acquisition-utility manifold of masked partial observations, where proximity reflects the expected utility of future acquisitions. Offline, training states are embedded and partitioned at each rollout timepoint; candidate future plans are scored within each partition using plug-in NOCT, and the best partition-level acquisition plans are cached. During inference, \OurMethod retrieves the cached plan via nearest-centroid lookup, enabling low-latency adaptive acquisitions. Experiments on synthetic and real-world healthcare datasets demonstrate that \texttt{\OurMethod} outperforms nonparametric, greedy, and RL-based alternatives, achieving higher accuracy at lower acquisition costs. Code is publicly available at \url{https://github.com/dzungdinhh/CAPE}.
\end{abstract}



\input{sections/1_intro}

\input{sections/2_related_works}
\input{sections/3_method}
\input{sections/4_experiments}
\input{sections/6_conclusion}
{
\small
\bibliographystyle{plainnat}
\bibliography{bibliography}
}

\clearpage
\appendix
\input{sections/7_appendix}

\end{document}

%% file: sections/1_intro.tex
\section{Introduction}
\label{intro}
In many real-world scenarios, \textit{the inference-time acquisition of information is a critical decision}. For example, in clinical longitudinal monitoring (a motivating application for this work), diagnostic decisions often rely on sequentially gathering information such as lab tests or imaging before predicting a patient's status. Each type of data acquisition comes at a cost of time, financial expense, or risk to the patient. In contrast, standard supervised machine learning pipelines typically assume that \emph{all} features are freely available at inference.

\begin{wrapfigure}{r}{0.50\textwidth}
    \centering
    \vspace{-15pt}
    \includegraphics[width=\linewidth]{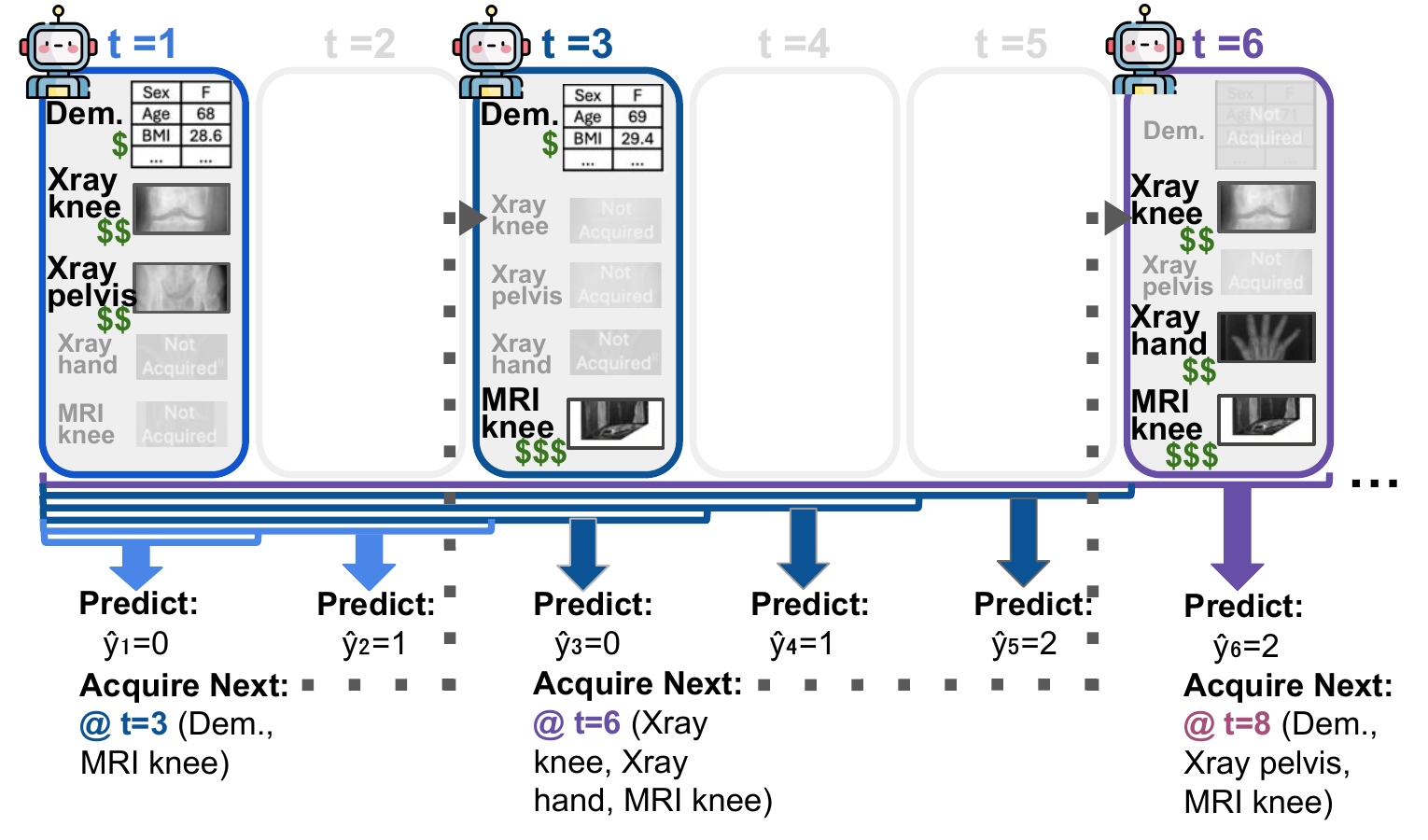}
    \caption{Illustration of clinical longitudinal AFA. At the current visit, an autonomous agent reviews previously collected and newly acquired data, predicts the patient's current status, and recommends a subset of examinations for a determined future time point. This process repeats until the agent recommends no further acquisition. \looseness=-1}
    \label{fig:AFA_example}
    \vspace{-10pt}
\end{wrapfigure}

In this work, we focus on \emph{longitudinal} Active Feature Acquisition (LAFA) \citep{saar2009active, kossen2022active}, where, during inference, a model is responsible for (i) sequentially deciding \emph{which features} to acquire and at \emph{what time}, balancing their utility against acquisition costs, and (ii) \emph{making predictions} with the partially observed features it has chosen to acquire.  In this setting, leveraging the temporal context is crucial for both selecting valuable feature acquisitions and making accurate predictions.
Fig.~\ref{fig:AFA_example} illustrates an example of LAFA by an autonomous agent in a clinical scenario. At the current visit, the agent 
(1) reviews previously collected data alongside any measurement acquired at the current visit, (2) predicts the patient's current status $\hat{y}_t$, and (3) decides when to schedule the next visit and which tests to acquire then (accounting for measurements' acquisition costs). Practical constraints such as time, cost, and resource limitations necessitate a judicious selection. 
Between visits, the agent continues to forecast the disease progression utilizing the previous acquisitions. 
This longitudinal AFA setting presents several challenges: deciding which features to acquire at each visit, facilitating early predictions for timely interventions, and accounting for temporal settings where missed acquisitions at earlier time points are permanently inaccessible. Although prior work has explored longitudinal AFA approaches, some methods~\citep{kossen2022active} make a single prediction across all timepoints, neglecting dynamic predictions that are critical for early detection and treatment-plan adjustment in clinical settings \citep{qin2024risk}. Furthermore, recent reinforcement learning (RL)-based methods~\citep{qin2024risk,nguyen2024active} are difficult to optimize in practice, while greedy strategies~\citep{gadgil2023estimating} make myopic decisions based on short-term gain and may skip early measurements that later prove critical but cannot be reacquired. This motivates nonmyopic methods that are practical to train and deploy. \looseness-1

We propose \texttt{\OurMethod} (\textbf{C}ontrastive \textbf{A}cquisition-\textbf{P}lan \textbf{E}mbeddings), a contrastive partitioning policy for longitudinal active feature acquisition. The key idea is to use oracle-based valuation to learn an AFA-based acquisition-utility manifold, where proximity between partial states reflects similarity in the expected utility of future feature-time acquisitions. States within the same local region, therefore, share acquisition-utility profiles, allowing candidate acquisition subsets to be reused with similar accuracy--cost effects. \OurMethod exploits this structure by partitioning the learned space, scoring future acquisition plans within each partition, and caching partition-level plans for fast deployment. This extends oracle-based AFA to longitudinal acquisition planning and turns expensive nonmyopic search into an offline policy-fitting procedure.

\textbf{Contributions}\quad \uline{First}, we use NOCT, an oracle-based plan objective that balances future predictive accuracy against feature-time acquisition cost, to supervise an acquisition-utility embedding space; to our knowledge, this is the first use of such a strategy in AFA. \uline{Second}, we develop a novel partitioning acquisition policy that amortizes planning over regions of this embedding space, enabling \OurMethod to select and revise future acquisition plans at deployment without expensive online plan scoring. \uline{Third}, across synthetic and real-world healthcare datasets, we show that \texttt{\OurMethod} outperforms a suite of nonparametric, greedy, and RL-based alternatives.



%% file: sections/2_related_works.tex
\section{Related Work}

\textbf{Active Feature Acquisition}\quad Prior works in AFA \citep{saar2009active, sheng2006feature} study the trade-off between prediction and feature acquisition cost in a non-longitudinal context. Recent works \citep{shim2018joint, yin2020reinforcement, janisch2020classification} frame AFA as an RL problem; however, they are challenging to train in practice due to the complicated state and action space with dynamically evolving dimensions. Other works \citep{li2021active} use surrogate generative models, which model multidimensional distributions, or employ greedy policies that ignore joint informativeness (some features are only informative in combination) \citep{covert2023learning, ma2018eddi, gong2019icebreaker}. To overcome these issues, \citet{ghosh2023difa} learn a differentiable acquisition policy by jointly optimizing selection and prediction models through end-to-end training. 

In addition, \citet{valancius2024acquisition} propose a nonparametric, oracle-based AFA method that uses training instances to evaluate candidate acquisitions. Our method is closest to this oracle-based line, but differs in how this information is represented and reused. A direct extension of \citet{valancius2024acquisition} would retrieve a fresh neighborhood of similar training instances in the native feature space and score many candidate plans online for each test state. This nearest-neighbor approach is computationally expensive, scaling linearly with the number of training instances, and may be more susceptible to the curse of dimensionality because neighborhoods are formed in the original high-dimensional feature space rather than in an acquisition-aware embedding. Instead, \OurMethod learns an acquisition-utility manifold in which local regions share similar future acquisition-utility profiles. It then partitions this space and fits a partitioning acquisition policy by scoring candidate future plans within each partition offline. At deployment, \OurMethod requires only a partition lookup before following the acquisition plan associated with the matched partition, enabling faster inference.



\textbf{Longitudinal Active Feature Acquisition}\quad Longitudinal AFA extends the standard AFA to include the temporal dimension.
In this setting, a policy must account for temporal constraints where past time points can no longer be accessed. As a result, a policy must decide which features to acquire and determine the optimal timing for each acquisition. For instance, \citet{zois2017active} derive dynamic‑programming and low‑complexity myopic policies for finite‑state Markov chains; ASAC \citep{yoon2019asac} uses an actor-critic approach to jointly train a feature selection and a prediction network. Additionally, \citet{nguyen2024active} use RL to optimize acquisition timing, but acquire all available features at the selected time point. Recently, \citet{qin2024risk} prioritize timely detection by allowing flexible follow-up intervals in continuous-time settings, ensuring that predictions are timely and accurate. 
However, RL-based approaches can be challenging to train due to issues such as sparse rewards, high-dimensional observation spaces, and complex temporal dependencies \citep{li2021active}. In contrast, \OurMethod uses the MDP formulation only as a formal lens for defining feasible future acquisition plans, while avoiding RL-based policy learning by learning an acquisition-utility manifold and fitting an offline partitioning acquisition policy that amortizes nonmyopic planning over local regions of the learned space. \looseness-1

%% file: sections/3_method.tex
\section{Methodology}
We propose \OurMethod, a method that learns a novel acquisition-utility manifold, an embedding space where proximity reflects the expected utility of future acquisitions. \OurMethod learns this manifold through two components: (1) a Nonmyopic Objective Cost-Tradeoff (NOCT) scores candidate future plans by their expected prediction loss and acquisition cost, producing acquisition-utility profiles that record which feature-time acquisitions consistently appear in the lowest-cost, highest-accuracy plans; (2) a contrastive objective supervised by these profiles pulls together instances with similar acquisition needs and pushes apart those with divergent ones.  \OurMethod then utilizes the manifold through a partitioning acquisition policy strategy that exploits nearby regions that share similar acquisition needs. At deployment, a new instance is embedded and mapped to its nearest partition to retrieve the respective acquisition plan, resulting in simple and efficient inference.

\subsection{Notation and Setup}
We consider longitudinal AFA over a dataset $\mathcal{D} = \{(x_i, y_i)\}_{i=1}^N$ of $N$ instances (e.g., patients). For each instance $i$, the covariates form a sequence of length $L$: $x_i = \{x_{i,t}\}_{t=1}^L$, where $t \in \{1, \ldots, L\}$ indexes the time step. At each time step $t$, $x_{i,t} \in \mathbb{R}^M$ contains $M$ (costly) feature measurements, but the policy may only acquire a subset at each visit \footnote{\texttt{\OurMethod} also supports multi-modal acquisitions where each modality yields a vector; this can be modeled by replacing scalar features with modality-specific vectors.}. We study a classification task with per-timepoint labels $y_{i,t} \in \{1, 2, \ldots, C\}$, where $C$ represents the number of classes. Following \citet{qin2024risk, yoon2019asac}, each feature $m$ has a time-invariant\footnote{\texttt{\OurMethod} easily extends to time-varying costs by replacing $c^m$ with $c^{m, t}$ (cost of modality $m$ at time $t$).} cost $c^m$. Finally, we use colon notation for subsequences; for example, $x_{1:t} = (x_1, \ldots, x_t)$ denotes measurements from timepoints $1$ to $t$ (inclusive).\looseness-1

At a timepoint $t$, let $o\subseteq \{(m,t')\mid m\in\{1,\ldots,M\}, t'\leq t\}$ denote the \textit{observed feature-time pairs}, and let $x_{i,o}$ be the corresponding partial observation. We assume a pre-trained conditioning classifier $\hat{y}$ that predicts the label at any target time $k$ from partial information, denoted as $\hat{y}_k(x_{i,o})$.

\subsection{Nonmyopic Objective/Cost Tradeoff (NOCT)}

We now introduce \texttt{\OurMethod} through a longitudinal oracle-based acquisition utility that provides a provable (see below) proxy for the MDP value of partially observed states. While prior oracle-based AFA objectives \citep{valancius2024acquisition} score candidates in the standard AFA setting, the introduced objective NOCT evaluates candidate feature-time plans in the longitudinal setting. Let $\mathcal{V}_{>t} = {\{(m, t') \mid m \in \{1, \ldots, M\}, \, t' \in \{t+1, \ldots, L\}\}}$ represent the set of candidate feature-time pairs that are still available to acquire after time $t$. At a partially observed state $(x_o, o, t)$, we have two options: (1) \emph{choose an acquisition plan} $v \subseteq  \mathcal{V}_{>t}$ (which specifies what to acquire in the future), or (2) \emph{terminate acquisition} (i.e., $v = \varnothing$) and make predictions for the remaining timepoints using only the currently observed information $x_o$.
The \textbf{n}onmyopic \textbf{o}bjective \textbf{c}ost-\textbf{t}radeoff (NOCT) determines the tradeoff of an acquisition plan of additional features $v \subseteq  \mathcal{V}_{>t}$ for a current partially known instance of $x_o$, where $o$ contains features from the past, up to time $\leq t$:
\begin{align}
\mathrm{NOCT}\bigl({x}_{o}, o, v\bigr)   \equiv \mathbb{E}_{y_{t+1:L}, x_v \mid {x}_{o}}[\ell (x_o \cup x_v, y, t)] + \alpha \!\!\! \sum_{(m, t') \in v} \!\!\! c^m,\label{eq:obj} 
\end{align}
where $\ell (x_o \cup x_v, y, t) \equiv \sum_{k=t+1}^L \mathrm{CE}\bigl(\hat{y}_k({x}_{o} \cup {x}_{v_{t+1:k}}),\,y_{k}\bigr) $ is the accumulated cross-entropy loss from $t{+}1$ to $L$ when acquiring the future feature-time pairs specified by $v$. 
Here, $x_{v_{t+1:k}}$ denotes the subset of acquired features in $v$ whose time indices lie in $\{t{+}1,\ldots,k\}$, i.e., the features that would be available by time $k$ under plan $v$, given the currently observed information $x_o$. Using training trajectories $(x_i,y_i)\in\mathcal{D}$, the expectation in Eq.~(\ref{eq:obj}) can be estimated by evaluating candidate plans on observed future outcomes during training.
Intuitively, Eq.~(\ref{eq:obj}) looks ahead at how acquiring a new set of feature-time pairs in $\mathcal{V}_{>t}$ is anticipated to reduce the classification loss from $t+1$ to $L$, while penalizing the acquisition cost. For simplicity, we refer to feature-time pairs as ``features''.  

\textbf{Connection to the optimal LAFA value}\quad
Although \OurMethod does not train an RL policy or use Bellman recursion, we relate NOCT to the optimal value of the sequential LAFA MDP.
Let $\mathbf{s}=(x_o,o,t)$ denote a partially observed state, and let $V^\star(\mathbf{s})$ be the optimal value of the corresponding finite-horizon LAFA Markov Decision Process (MDP), where rewards are negative prediction loss and acquisition cost. The full MDP formulation and proof are given in Appx.~\ref{appendex:theorem}.

\begin{theorem}[NOCT lower bound]
\label{thm:lowerbound}
Let $\mathbf{s}=(x_o,o,t)$ be a partially observed state in the finite-horizon LAFA MDP with fixed predictor $\hat{y}$ and nonnegative acquisition costs. 
\begin{equation}
\text{If}\ J_{\mathrm{NOCT}}(\mathbf{s})
=
-\min_{v\subseteq \mathcal{V}_{>t}} \mathrm{NOCT}(x_o,o,v),\  \text{then}\  J_{\mathrm{NOCT}}(\mathbf{s})\le V^\star(\mathbf{s}).
\label{eq:theorem_eq}
\end{equation}
\end{theorem}

Thus, $-\mathrm{NOCT}$ lower bounds the optimal longitudinal AFA value and provides a principled surrogate for evaluating future acquisition plans. This result motivates NOCT as an objective for comparing future acquisition plans, but does not provide approximation or regret guarantees for the practical \OurMethod policy. In practice, \OurMethod does not solve Eq.~\eqref{eq:theorem_eq} exactly at deployment. Instead, it estimates NOCT from training trajectories to learn an acquisition-utility manifold and fits a partitioning acquisition policy. The following subsections describe how this approximation yields an efficient deployment procedure by amortizing nonmyopic planning over partitioned regions of the learned space.\looseness-1
\subsection{Acquisition-Aware Contrastive Representations}

\label{subsec:nonparam}

We learn an embedding $g_\phi$ that maps partial states according to their NOCT-guided future acquisition-utility profiles, using a contrastive objective. Intuitively, $g_\phi$ is trained so that nearby states benefit similarly from future feature acquisitions. This structure underlies the efficacy of the partitioning policy in Sec.~\ref{subsec:cache}, which tailors acquisition strategies to concentrated local regions of the learned manifold that, by construction, contain partial states whose candidate acquisition subsets induce similar accuracy--cost tradeoffs. Thus, the embedding defines an AFA-based acquisition-utility manifold, rather than a generic similarity space.

\textbf{Plug-in NOCT}\quad For \emph{training instances}, $(x_i, y_i) \in \mathcal{D}$, the direct estimate of $\mathrm{NOCT}\bigl({x}_{i,o}, o, v\bigr)$ (Eq.~(\ref{eq:obj})) is a look-ahead plug-in sample approach, which we use to learn our manifold:
\begin{equation}
    \scalebox{0.9}{$\displaystyle
        \widehat{\mathrm{NOCT}}_{pl}\bigl({x}_{i,o}, o, v\bigr) =\ell (x_{i,o} \cup x_{i,v}, y_i, t) + \alpha \!\!\! \sum_{(m, t') \in v} \!\!\! c^m.
    $}
    \label{eq:plugobj}
\end{equation}
\textbf{Acquisition-Utility Profiles}\quad For each partial state $(x_{i,o},o,t)$, we want a summary of which future measurements are expected to be useful. Here, our key idea is that if certain feature-time acquisitions repeatedly appear in low-NOCT future plans, then those acquisitions characterize the utility profile of the current partial state. We represent this profile by a distribution $\zeta_{i,o}$ over future acquisitions $(m,t')\in \mathcal{V}_{>t}$ and termination $\varnothing$. 

To construct $\zeta_{i,o}$ for a training instance, we rank candidate future plans $v\subseteq \mathcal{V}_{>t}$ using the plug-in objective $\widehat{\mathrm{NOCT}}_{\mathrm{pl}}(x_{i,o},o,v)$ and take the top-$r$ plans $\hat{v}_{i,o}^{(1)}, \ldots, \hat{v}_{i,o}^{(r)}$ with the smallest values. The profile $\zeta_{i,o}(m,t')$ is then given by a weighted frequency of how often $(m,t')$ appears in these top plans, with weights based on their plug-in NOCT scores. Thus, $\zeta_{i,o}$ records which future measurements are consistently suggested by the best plans, and we use similarities between these profiles to supervise the contrastive embedding. We provide the formal $\zeta_{i,o}$ construction in Appx.~\ref{appendex:future_dist}. 

\begin{figure}[!t]
\centering
    \begin{subfigure}[t]{0.45\linewidth}
        \centering
        \includegraphics[width=\linewidth]{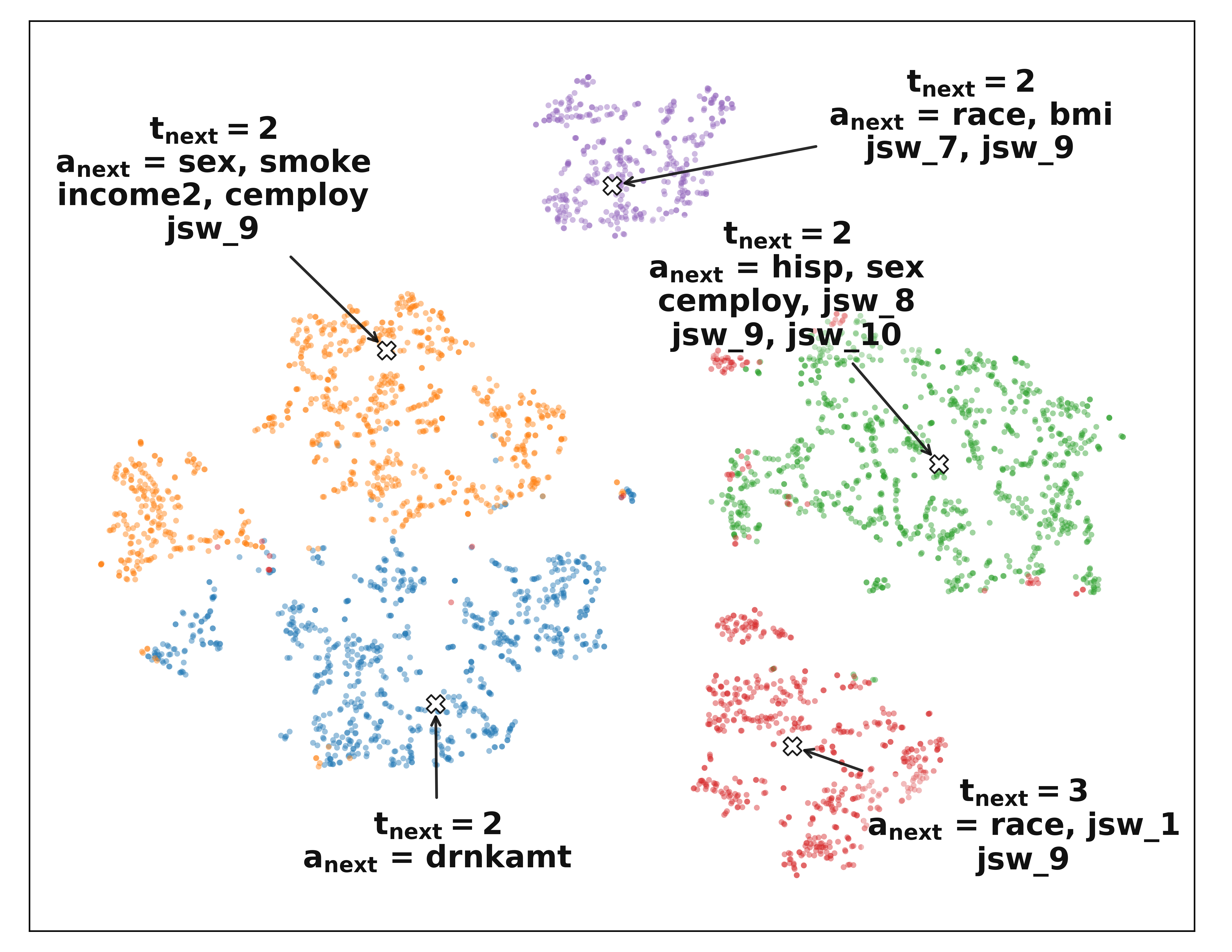}
        \caption{Acquisition regions at time $t=1$.}
        \label{fig:embedding_regions}
    \end{subfigure}
    \hspace{1cm} 
    \begin{subfigure}[t]{0.45\linewidth}
        \centering
        \includegraphics[width=\linewidth]{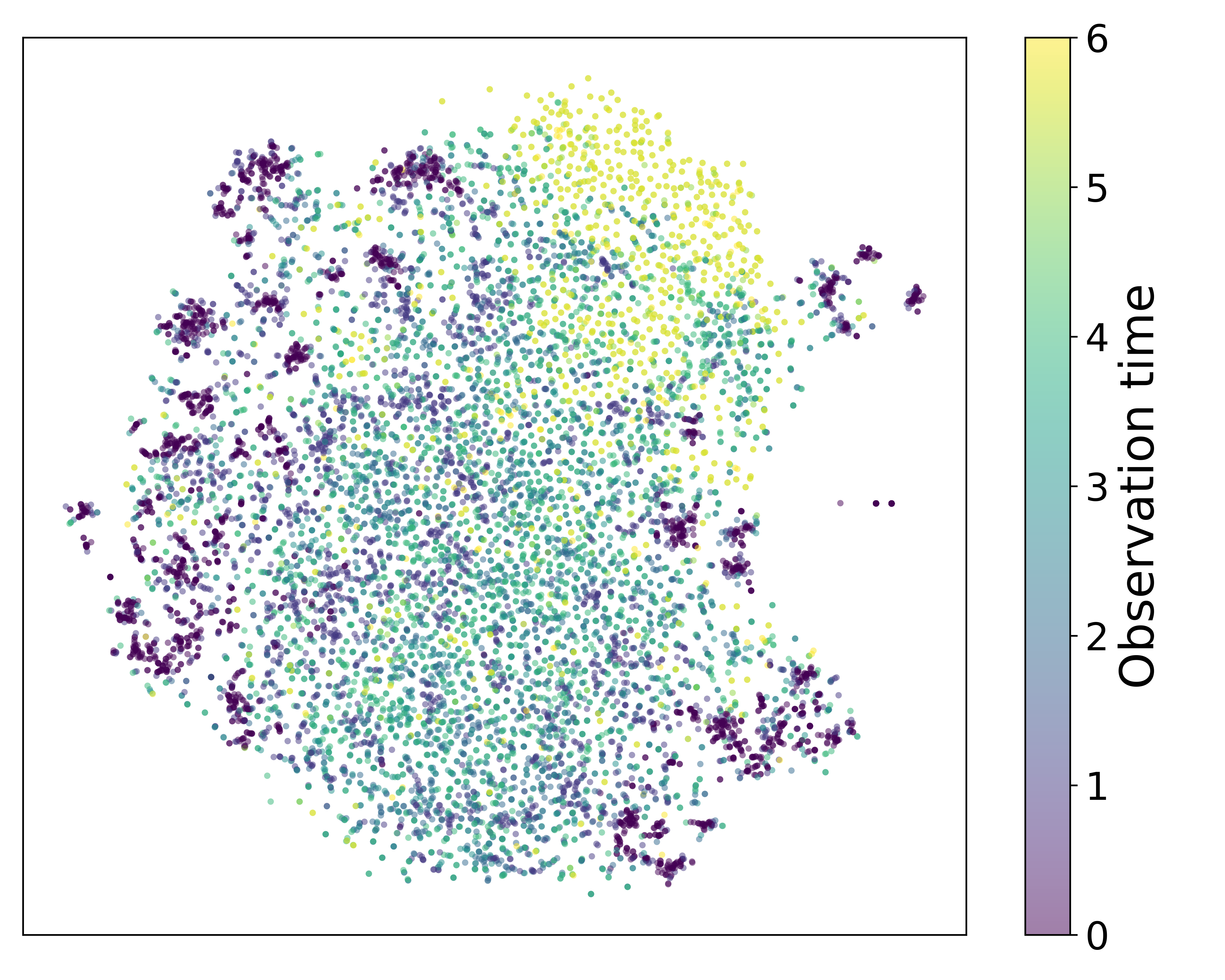}
        \caption{t-SNE colored by current timepoint.}
        \label{fig:embedding_tsne_time}
    \end{subfigure}
    \caption{
    Visualizations of the acquisition-aware embedding on the KLG task.
    \textbf{(a)} A t-SNE visualization of KLG patient embeddings at rollout time $t=1$. Points denote training states, crosses mark region centers, and annotations show the region-specific next acquisition, where $t_{\mathrm{next}}$ denotes the next visit and $a_{\mathrm{next}}$ denotes the selected measurements. Each color corresponds to a local region of the learned space that shares a cached acquisition decision.
    \textbf{(b)} A t-SNE visualization of rollout-state embeddings colored by current observation timepoint. The learned representation exhibits temporal structure, suggesting that $g_\phi$ captures differences in acquisition context across rollout stages. 
    }
    \label{fig:embedding_diagnostic}
\end{figure}

\textbf{Contrastive Learning}\quad We learn a manifold that maps partially observed features into an embedding space where distance reflects similarity of acquisition-utility profiles. Formally, we learn an embedding network $g_\phi(x_{i,o}, t) \in \mathbb{R}^E$ such that two partial states $(x_{i,o},o,t)$ and $(x_{j,o},o,t)$ are mapped nearby when they share similar top future acquisitions, meaning that the same acquisition decisions are expected to produce similar accuracy gains under comparable cost budgets. Conversely, instances with divergent acquisition-utility profiles are pushed away in $\mathbb{R}^E$. Fig.~\ref{fig:embedding_regions} illustrates this structure, where the local regions in the learned space correspond to different reusable next-acquisition decisions. \looseness-1

To achieve this, we define a similarity score between instances based on their future-candidate distributions $\zeta_{i, o}$ and $\zeta_{j, o}$, using the exponential of their negative \emph{Jensen-Shannon ($\mathrm{JS}$) divergence}:
\begin{equation}
      \texttt{sim}_{ij}
    = \exp\!\bigl(-\beta\,\mathrm{JS}(\zeta_{i, o}, \zeta_{j, o})\bigr),
    \label{eq:js}
\end{equation}
$\beta > 0$ controls how strongly divergence affects similarity. Letting $\delta_{ij} = {\bigl\lVert g_\phi \bigl(x_{i,o}\bigr) - g_\phi\bigl(x_{j, o}\bigr)\bigr\rVert^2}$ be the embedding distance, we minimize a contrastive loss \citep{chopra2005learning}:
\begin{equation}
\label{eq:ebmedding}
    \scalebox{0.87}{$\displaystyle
    \mathcal{L}_{\mathrm{emb}} = \frac{1}{2} \Bigl(\texttt{sim}_{ij} \cdot \delta_{ij} + 
    (1 - \texttt{sim}_{ij}) \cdot \max \bigl(0,\,\gamma - \delta_{ij}\bigr)^2\Bigr)$},
\end{equation} 
where $\gamma > 0$ is the margin parameter. Minimizing $\mathcal{L}_{\mathrm{emb}}$ encourages $g_\phi$ to create an embedding space where distance reflects the similarity of future-candidate distributions. 

Fig.~\ref{fig:embedding_tsne_time} shows that rollout states exhibit clear temporal structure in the embedding space. This motivates using the embedding not merely for retrieval, but to define local policy regions within the learned acquisition-utility manifold, where a temporal partition-level acquisition plan can be reused across states with similar expected utility for future acquisitions.

\subsection{Partitioning Acquisition Policy Construction}
\label{subsec:cache}
In the acquisition-utility manifold, nearby states are expected to exhibit similar useful future-acquisition behavior. We use this structure to make nonmyopic planning efficient at inference time: we partition concentrated regions of the learned embedding space at each time step $t$, fit a representative acquisition plan within each partition, and reuse that plan for nearby states.


\textbf{Embedding-based Partitioning}\quad At time $t$, let $x_{i,o}$ denote the partially observed training instance for sample $i$. Using the embedding network $g_\phi$, we compute acquisition-aware embeddings $e_{i,o}^t = g_\phi(x_{i,o}, t)$. We then cluster, e.g., using K-Means,  $\{e_{i,o}^t\}_{i=1}^{n}$ \emph{to partition the learned acquisition-utility manifold into $K$ concentrated regions} $C_{1,t},\ldots,C_{K,t}$ with centroids $\mu_{1,t},\ldots,\mu_{K,t}$. Because $g_\phi$ is trained so that distances reflect similarity in acquisition-utility profiles, each region serves as a policy partition whose members are expected to share similar future acquisition behavior.

\textbf{Partitioning Acquisition Policy}\quad For each partition $C_{k,t}$, we sample a partition-level candidate set $\mathcal{P}_{k,t}\subseteq 2^{\mathcal V_{>t}}$, including future acquisition plans and termination plan $\varnothing$. For each candidate plan $v\in\mathcal{P}_{k,t}$, we compute its average plug-in NOCT score over the partition:
\begin{equation}
\widehat{\mathrm{NOCT}}_{k,t}(v)
=
\frac{1}{|C_{k,t}|}
\sum_{i\in C_{k,t}}
\widehat{\mathrm{NOCT}}_{\mathrm{pl}}
\bigl(x_{i,o},o,v\bigr).
\label{eq:cluster_noct}
\end{equation}
Intuitively, lower values of $\widehat{\mathrm{NOCT}}_{k,t}(v)$ indicate better accuracy-cost tradeoffs for the partition. We therefore cache the top acquisition plan for partition $k$ as:
\begin{equation}
\hat{v}_{k,t}^{*} = \argmin_{v\in\mathcal{P}_{k,t}} \widehat{\mathrm{NOCT}}_{k,t}(v)
\label{eq:cached_plan}
\end{equation}
where $\hat{v}^{*}_{k,t}$ is the candidate plan with the smallest partition-level plug-in NOCT scores. The cache stores the embeddings $\{e_{i}^t\}_{i=1}^{n}$, centroids $\{\mu_{k,t}\}_{k=1}^K$, cached plans $\{\hat{v}^{*}_{k,t}\}_{k=1}^K$, and their scores.


Policy-cache construction proceeds as a rollout over the training set. After the partition plans are cached at time $t$, each training instance is assigned to its respective policy partition, the corresponding cached plan is selected, and only the next immediate (temporally) acquisitions from that plan are executed (e.g., see Fig.~\ref{fig:embedding_regions}). Each instance’s observed set $o$ is then updated, and the instance advances to the corresponding next timepoint. Policy fitting then proceeds to time $t+1$.

\subsection{Policy Deployment}
\begin{wrapfigure}{r}{0.50\linewidth}
    \vspace{-2.3em}
    \begin{minipage}{0.50\textwidth}
        \begin{algorithm}[H]
            \small
            \caption{\texttt{\OurMethod} Deployment 
            \label{alg:cached_deploy}}
            \begin{algorithmic}[1]
                \Require Predictor $\hat y_k$, embedding $g_\phi$, cached embeddings $\{e_{i}^t\}$, cached centroids $\{\mu_{k,t}\}$, cached plans $\{\hat{v}^{*}_{k,t}\}$.
                \State Initialize: \texttt{predictions} $\gets [\,]$; \texttt{terminate} $\gets$ \texttt{false}; $t\gets 0$; $o\gets\varnothing$
                \While{\textbf{not} \texttt{terminate} \textbf{and} $t < L$}
                    \State $e_o^t \gets g_\phi(x_o,t)$
                    \State Find nearest cached centroid $\mu_{k,t}$ to $e_o^t$
                    \State Select cached plan $\hat{v}^{*}_{k,t}$ for centroid $\mu_{k,t}$
                    \If{$\hat v^{*}=\varnothing$}
                        \State $t_{\text{next}}\gets L$; \texttt{terminate} $\gets$ \texttt{true}
                    \Else
                        \State $t_{\text{next}}\gets\min\{\,t'\mid(m,t')\in\hat {v}^{*}_{k,t}\,\}$
                        \State $o \gets o \cup \{(m,t_{\text{next}})\mid(m,t_{\text{next}})\in\hat{v}^{*}_{k,t}\}$
                    \EndIf
                    \For{$t'=t+1,\ldots,t_{\text{next}}$}
                        \State append $\hat y_{t'}(x_{o_{1:t'}})$ to \texttt{predictions}
                    \EndFor
                    \State $t\gets t_{\text{next}}$
                \EndWhile
                \State \Return \texttt{predictions}
            \end{algorithmic}
        \end{algorithm}
    \end{minipage}
    \vspace{-1.5em}
\end{wrapfigure}

At partitioning policy deployment, \OurMethod uses the learned acquisition-utility manifold to identify which policy partition region best matches the current partial observation  (Alg.~\ref{alg:cached_deploy}). Given a test state $(x_o,o,t)$, we compute the associated embedding $e_o^t=g_\phi(x_o,t)$ and map it to the nearest cached centroid $\mu_{k,t}$ with $k=\arg\min_{k\in\{1,\ldots,K\}}\|e_o^t-\mu_{k,t}\|_2^2$. Intuitively, the cached plan $\hat{v}^{*}_{k,t}$ is the plan that achieved the best average NOCT tradeoff for similar training states at the same local acquisition-utility region. \looseness-1

The cached plan $\hat{v}^{*}_{k,t}$ is a future acquisition schedule with potential rather than a commitment. This keeps the policy adaptive: each new observation can change the utility of remaining acquisitions, so rather than following the full plan rigidly, \texttt{\OurMethod} reassesses at each step. If $\hat{v}^{*}_{k,t}=\varnothing$, then the cached region suggests that further acquisition is not worth its cost and \OurMethod terminates, predicting remaining timepoints from current information. Otherwise, \OurMethod executes only the earliest acquisitions in $\hat{v}^{*}_{k,t}$, which acquire features at the next earliest occurring timepoint $t_{\text{next}}$ in $\hat{v}^{*}_{k,t}$.
Rather than following the rest of the previously selected plan, \OurMethod updates $o$, advances to the next decision timepoint, $t_{\text{next}}$, and repeats the cached lookup.
This mirrors the clinical intuition of ordering the next informative test, observing the result, and then reassessing.



%% file: sections/4_experiments.tex
\section{Experiments}
\label{sec:experiment}
\subsection{Datasets}
\textbf{Synthetic}\quad Inspired by \citet{kossen2022active}, we construct a synthetic dataset of $N=8{,}000$ samples with $L=10$ non-i.i.d. timepoints. Each timepoint contains two features: \textit{digit} and \textit{counter}. Labels are the cumulative sum of the digits when the counter values are zero. Specifically, the \textit{digit} feature is a sequence of uniformly random $L$ integers from $\{0, 1, 2\}$, e.g., $1010220110$. The \textit{counter} feature is created by concatenating countdown sequences that begin with randomly chosen starting values from $\{0, 1, 2\}$ and truncating to a fixed length of $L$. For example, starting values $2$, $1$, $2$, and $1$ yield $210$, $10$, $210$, $10$, and the final counter feature is their concatenation $2101021010$ (of length $L$). Unlike \citet{kossen2022active}, we assign labels at every timestep: 
$y_t=\sum_{t'=1}^{t}\mathrm{digit}_{t'}\mathbb{I}\{\mathrm{counter}_{t'}=0\}$; for example, label $0011333444$ in the case of $\genfrac{}{}{0pt}{}{\mathrm{counter:}}{\mathrm{digit:}}  
 \genfrac{}{}{0pt}{}{21\underline{0}1\underline{0}21\underline{0}1\underline{0}}{10\mathbf{1}0\mathbf{2}20\mathbf{1}1\mathbf{0}}$. The policy should begin with acquiring the digit and counter at 
$t=1$, then acquiring the digit at the next anticipated timestep where $\mathrm{counter}=0$, then acquiring the \textit{counter} again to locate the next $\mathrm{counter}=0$ timestep, and repeat. Following \citet{kossen2022active}, we split data into train/val/test ($70\%/15\%/15\%$) \looseness-1.

\textbf{OAI}\quad The Osteoarthritis Initiative (OAI; \url{https://nda.nih.gov/oai/}) contains $N=4{,}796$ patients with left and right knee records followed for up to $96$ months. We use $27$ visit-level features, including tabular features from~\citet{chen2024unified} and joint space width, across $L=7$ visits. We predict KLG ($0 \sim 4$)~\citep{kellgren1957radiological} and WOMAC pain ($0 \sim 20$)~\citep{mcconnell2001western}; following prior work~\citep{chen2024unified,nguyen2024active}, we merge KLG $0/1$, binarize WOMAC pain at $5$, and use a $50\%/12.5\%/37.5\%$ train/val/test split.

\textbf{ILIADD}\quad The Intensive Longitudinal Investigation of Alternative Diagnostic Dimensions (ILIADD) dataset~\citep{Ringwald2025DoYF} is a remote mental-health study with $544$ participants, $34$ demographic/HiTOP-SR-derived features, and up to eight EMA surveys per day. We use length-$L=10$ windows to predict negative affect, yielding $N=40{,}279$ train/val/test windows ($69.5\%/15.25\%/15.25\%$).\looseness-1

\textbf{Supplemental Materials}\quad Please see Appx.~\ref{sec:addl_results} for additional results and datasets.

\subsection{Baselines}
\emph{We {expand the comparisons} in recent work} of \citet{qin2024risk} in longitudinal AFA  by including \emph{three additional nonuniform-feature-cost-capable state-of-the-art AFA methods}, {DIME}~\citep{gadgil2023estimating}, {DiFA} \citep{ghosh2023difa}, {ACO} \citep{valancius2024acquisition}, together with the RL method {ASAC}~\citep{yoon2019asac} from \citet{qin2024risk}.  
We consider 6 baselines in total: 1) \textbf{ASAC}~\citep{yoon2019asac}: an actor-critic method to jointly train a feature selector and a predictor network; 2) \textbf{RAS}~\citep{qin2024risk}: an RL method for continuous-time longitudinal acquisition; 3) \textbf{AS} \citep{qin2024risk}: RAS with a constant acquisition interval; 4) \textbf{DIME}~\citep{gadgil2023estimating}: sequential selection by marginal utility; 5) \textbf{DiFA} \citep{ghosh2023difa}: a Gumbel-softmax-based differentiable policy; 6) \textbf{ACO} \citep{valancius2024acquisition}: an oracle-based acquisition policy. For each baseline, \emph{we perform a comparable hyperparameter search on the validation set using the authors' original code}; full details are in Appx.~\ref{sec:baseline}.

We adapt DIME and DiFA for longitudinal data by restricting acquisitions to either the current or future time points. Note that DIME and DiFA acquire features one at a time within a timestep, while \texttt{\OurMethod} and other baselines select all desired features at $t$ simultaneously before advancing.
This gives DIME and DiFA additional within-timepoint adaptivity: after observing each acquired feature value, they can decide which feature to acquire next at the current timepoint or whether to advance to a later timepoint.\looseness-1

\subsection{Implementation}
\label{sec:implementation}

\textbf{Data Availability and Missingness}\quad Unlike standard AFA methods that assume fully observed data \citep{rahbar2025survey, shim2018joint, nan2015feature}, our work is general and handles missingness. This is done by excluding missing features from candidate sets and pretraining the prediction network $\hat{y}_k$ and embedding network $g_\phi$ with random input-feature dropout.

\textbf{Prediction Network Implementation}\quad  
At each time $t$, an MLP predicts outcomes $\hat{y}_k(x_{o_{1:t}})$ for $k \geq t$ from the observed features $x_{o_{1:t}}$\footnote{Inherently missing and unacquired features are masked out.} and the target time indicator $k$, using only observations available through $t$. We train the network with random input dropout to improve robustness to varying missingness patterns.


\textbf{\OurMethod Implementation.}\quad
We train the MLP predictor $\hat{y}_k$ and then train the MLP embedding network $g_\phi$. We set $\beta=\gamma=1$, $K\in\{5, 10, 20, 50, 100 \}$, and embedding dimension $E=32$ (Synthetic) and $E=64$ (OAI/ILIADD). We utilize Adam~\citep{kingma2014adam} with a learning rate of $10^{-3}$. During the partitioning acquisition policy construction, we sample $|\mathcal{P}|=5{,}000$ acquisition plans.



\subsection{Performance-Cost Results}

\begin{figure*}[t] 
  \centering
  \includegraphics[width=\textwidth]{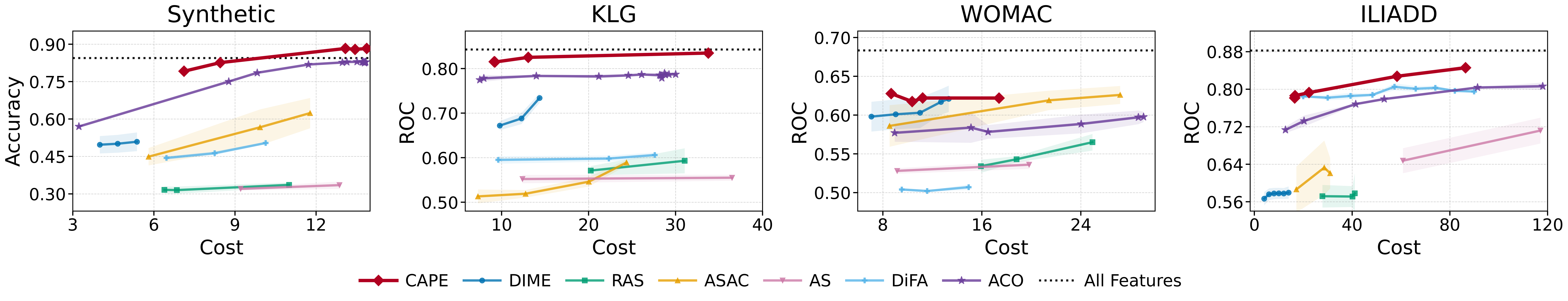}
  \vspace{-1.7em}
  \caption{Performance/cost of models across various average acquisition costs (budgets), with standard error across 5 runs. Following \citet{kossen2022active}, we report accuracy instead of AUROC for synthetic data. The dotted horizontal line shows performance using all features. \OurMethod consistently yields higher predictive performance at comparable or lower acquisition costs.}
  \label{fig:performance}
  \vspace{-8pt}
\end{figure*}

Fig.~\ref{fig:performance} shows performance against acquisition cost. For each panel, the x-axis represents different average acquisition costs per instance, and the y-axis represents the predictive performance. Higher budgets allow more acquisitions and typically improve performance. Performance-cost points are obtained by sweeping the cost-control parameter $\alpha$ (in Eq.~(\ref{eq:obj})) for \OurMethod and the corresponding cost-control hyperparameters for each baseline. See Appx.~\ref{appendex:experiment_setup} for details on optimal model selection for target budgets and dataset-specific feature costs. Following  \citep{kossen2022active, qin2024risk}, we report the synthetic dataset accuracy and real-world healthcare datasets AUROC. \looseness-1

\textbf{Results}\quad \OurMethod achieves the strongest overall performance--cost tradeoff across all datasets, with the best or near-best accuracy/AUROC across the available acquisition budgets. \OurMethod remains above competing acquisition policies over most budgets on Synthetic, WOMAC, KLG, and ILIADD. These results suggest that the learned acquisition-utility manifold provides reusable acquisition decisions that generalize across similar partial states. DIME and ACO perform well in some regimes but are less consistent across datasets and budgets. RAS, ASAC, AS, and DiFA often improve as acquisition cost increases, but generally remain below \OurMethod and sometimes exhibit notably weaker performance in the low-cost regime. This regime is especially important, as a useful policy should identify high-value measurements before the budget becomes large enough to approximate the all-features setting. Additional results on Average Precision are in Appx.~\ref{appendex:prauc}.

\begin{wrapfigure}{r}{0.45\textwidth}
    \centering
    \vspace{-12pt}
    \includegraphics[trim=0pt 0pt 0cm 0pt, clip, width=\linewidth]{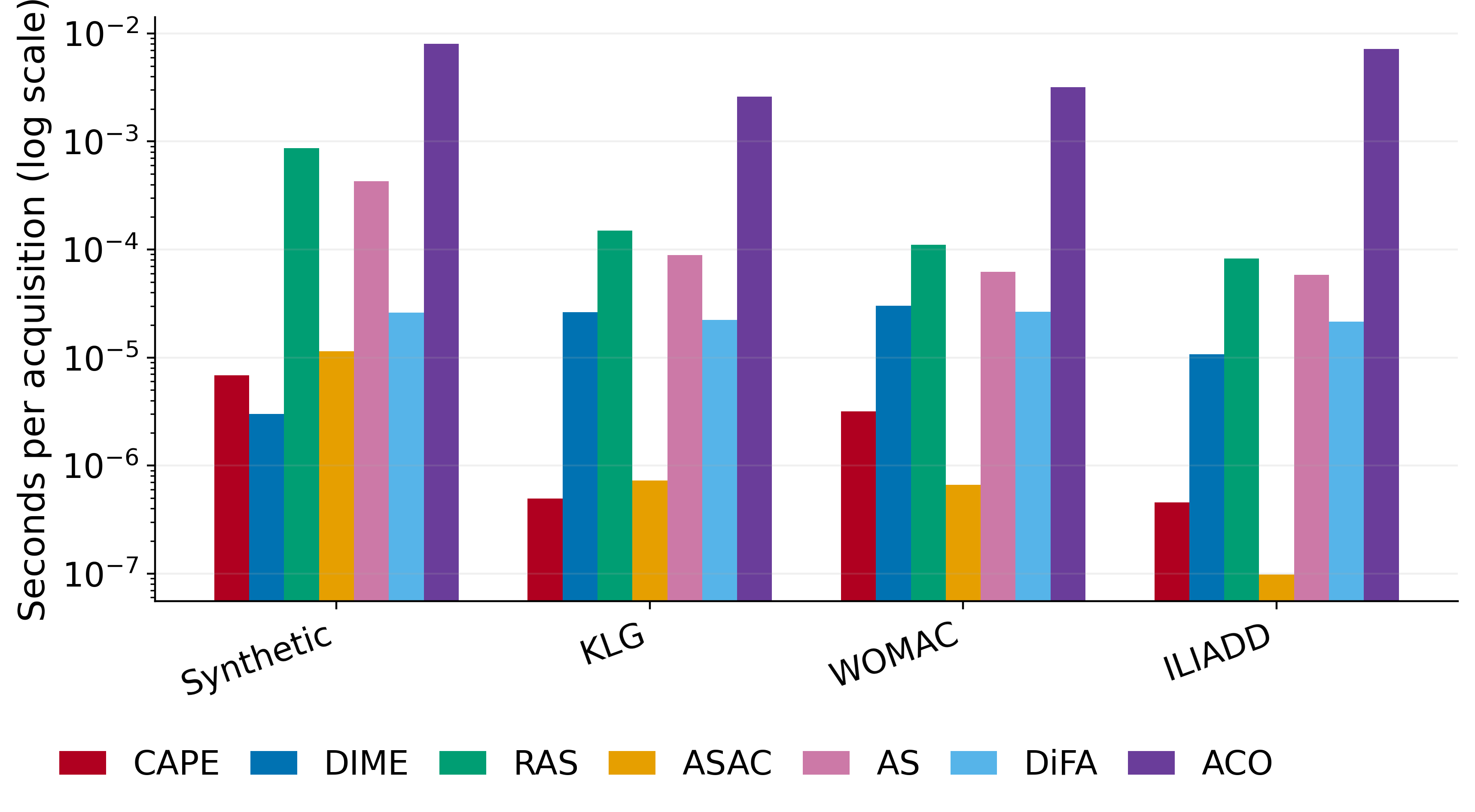}
    \vspace{-1.7em}
    \caption{Inference latency per acquisition.}
    \label{fig:timing_eval}
    \vspace{-10pt}
\end{wrapfigure}
\textbf{Inference Time}\quad \OurMethod provides a favorable tradeoff between predictive performance and deployment efficiency. Across datasets, it is the most consistent leader in accuracy/AUROC as acquisition cost increases, as shown in Fig.~\ref{fig:performance}.  
Importantly, this performance does not come at the expense of inference-time overhead (see Fig.~\ref{fig:timing_eval}): \OurMethod is also among the fastest methods per acquisition, often requiring only a lightweight partition lookup at test time. In contrast, ACO, the nearest-neighbor oracle-based baseline, is the next strongest competitor on the cost-benefit tradeoff, but it is also among the slowest methods, often orders of magnitude slower than \OurMethod. ASAC sits at the opposite extreme: fast but underperforming on the cost-benefit tradeoff. Thus, the figures show that \OurMethod does not sacrifice speed/accuracy for the other--\textbf{it occupies the strongest region of the joint accuracy–latency frontier}.\looseness-1

\begin{table*}[t]
\captionsetup{skip=2pt}
\caption{Ablation comparing acquisition-aware embeddings $g_\phi(\cdot)$ used by \texttt{CAPE} against partitioning directly in the raw feature-value space. We report the mean $\pm$ standard errors across five runs.}
\label{tab:embedding_ablation}
\begin{center}
\begin{sc}
\resizebox{1\linewidth}{!}{%
\begin{tabular}{c
                cc @{\quad}
                cc @{\quad}
                cc @{\quad}
                cc}
\toprule
\multirow{2}{*}{Method}
  & \multicolumn{2}{c}{Synthetic}
  & \multicolumn{2}{c}{KLG}
  & \multicolumn{2}{c}{WOMAC}
  & \multicolumn{2}{c}{ILIADD} \\
\cmidrule(lr){2-3} \cmidrule(lr){4-5} \cmidrule(lr){6-7} \cmidrule(lr){8-9}
 & Acc. & cost
 & AUROC & cost
 & AUROC & cost
 & AUROC & cost \\
\midrule
\multirow{2}{*}{\makecell{Feature\\Partitioning}}
 & $0.774\pm0.000$ & $7.178\pm0.005$
 & $0.804\pm0.002$ & $9.912\pm0.046$
 & $0.618\pm0.005$ & $8.461\pm0.055$
 & $0.762\pm0.009$ & $17.283\pm0.985$ \\
 & $0.866\pm0.000$ & $13.932\pm0.004$
 & $0.823\pm0.001$ & $34.931\pm0.053$
 & $0.610\pm0.004$ & $17.073\pm0.204$
 & $0.822\pm0.004$ & $81.634\pm1.818$ \\
\midrule
\multirow{2}{*}{\texttt{CAPE}}
 & $\mathbf{0.792}\pm\mathbf{0.000}$ & $7.114\pm0.005$
 & $\mathbf{0.815}\pm\mathbf{0.004}$ & $9.147\pm0.082$
 & $\mathbf{0.628}\pm\mathbf{0.004}$ & $8.673\pm0.275$
 & $\mathbf{0.786}\pm\mathbf{0.001}$ & $16.672\pm1.717$ \\
 & $\mathbf{0.883}\pm\mathbf{0.000}$ & $13.877\pm0.008$
 & $\mathbf{0.835}\pm\mathbf{0.003}$ & $33.758\pm1.271$
 & $\mathbf{0.622}\pm\mathbf{0.004}$ & $17.395\pm0.204$
 & $\mathbf{0.846}\pm\mathbf{0.005}$ & $86.450\pm1.818$ \\
\bottomrule
\end{tabular}%
}
\end{sc}
\end{center}
\end{table*}

\subsection{Ablation of the Acquisition-Aware Embedding}
We evaluate the effectiveness of the learned acquisition-aware representation $g_\phi(\cdot)$ for constructing acquisition regions. To isolate this effect, we keep the same partitioning policy construction and replace only the representation space used to partition observation states. Specifically, the raw feature-value baseline partitions states using the observed feature values $x_o$, comparing two observations $x_{i,o}$ and $x_{j,o}$ by $d(x_{i,o},x_{j,o})=\|x_{i,o}-x_{j,o}\|_{2}^2$.
The result in Table~\ref{tab:embedding_ablation} shows that this learned representation 
\begin{wrapfigure}{r}{0.58\textwidth}
  \centering
  \includegraphics[width=\linewidth]
    {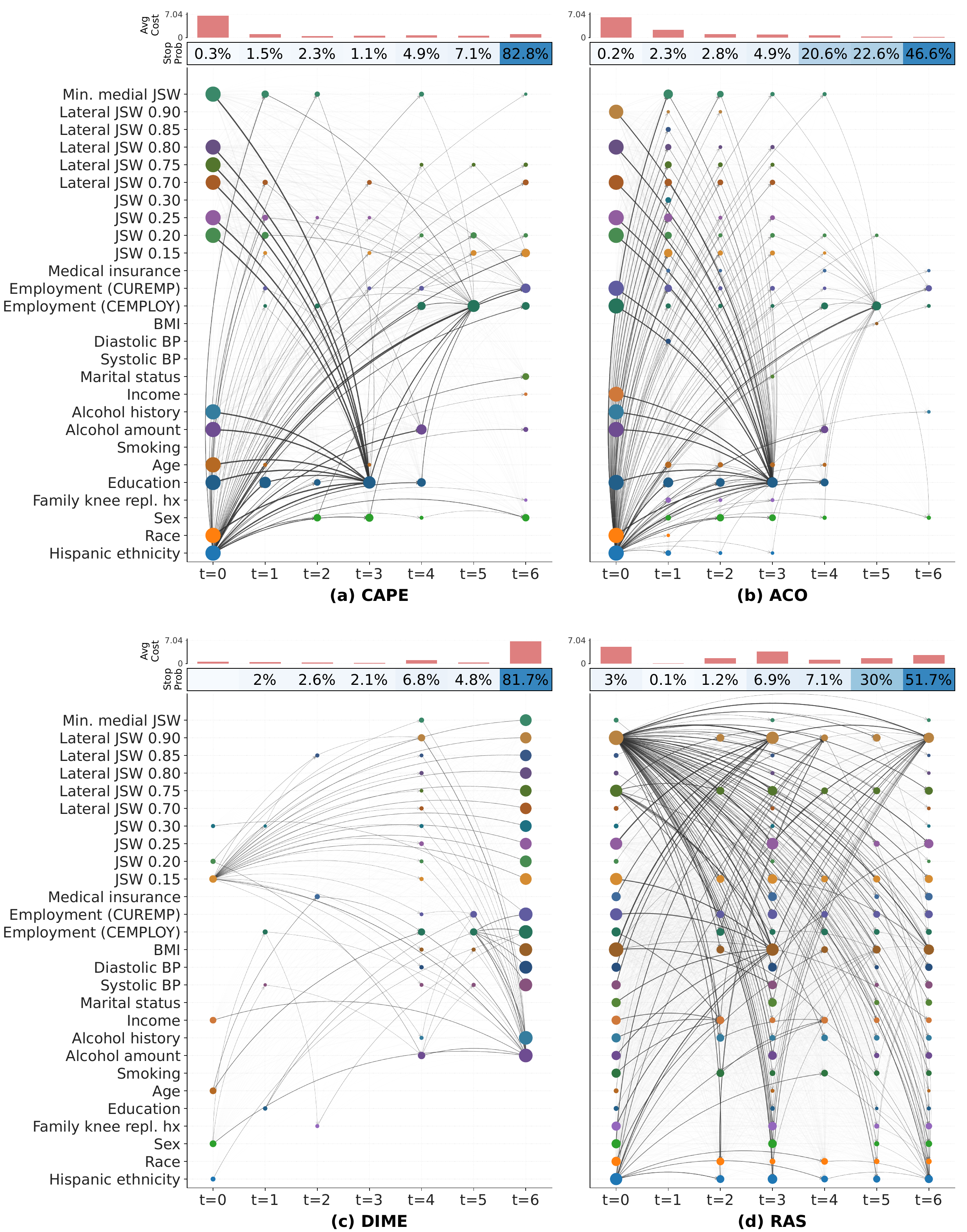}
  \vspace{-0.5em}
  \caption{Acquisition dynamics on the WOMAC test set.
  For each method, \textbf{top} shows the mean acquisition
  cost per sample at each timepoint, and \textbf{middle}
  shows the proportion of samples terminating at each
  timepoint. The \textbf{bottom} graph shows feature--time
  acquisitions, with node size indicating acquisition
  frequency. A directed edge from $(m,t)$ to $(m',t')$,
  where $t'>t$, indicates cross-time co-acquisition:
  samples acquiring $(m,t)$ also acquire $(m',t')$ later.
  Mean total acquisition costs per sample:
  $11.22$ (\OurMethod), $11.89$ (ACO),
  $9.95$ (DIME), and $15.93$ (RAS).
  \vspace{-20pt}}
  \label{fig:qual_dynamics}
\end{wrapfigure}
\noindent is crucial for effective partitioning. Across all  datasets, partitioning in the learned embedding space improves predictive performance over partitioning directly in the raw feature-value space, while incurring comparable acquisition cost, which suggests that the improvement is not due to acquiring more information, but rather to selecting more useful acquisitions within the same policy-construction framework. Finally, the fact that raw feature-space partitioning often remains competitive with ACO provides further evidence for the underlying partition-based policy principle, while the additional gains from $g_\phi(\cdot)$ demonstrate the value of learning an acquisition-aware geometry.

\subsection{Acquisition Trajectory Analysis}
Fig.~\ref{fig:qual_dynamics} illustrates distinct acquisition dynamics on the WOMAC prediction task. \OurMethod{} combines early acquisition of demographic and clinical information with sustained follow-up through sparse, dynamic, feature-specific updates. Several joint-space-width measurements are selected for more patients at baseline than at later visits, where acquisition costs remain low. ACO similarly emphasizes baseline acquisition, while \OurMethod{} achieves this pattern with faster inference using NOCT-guided cached plans. 
In contrast, RAS repeatedly acquires a broad, relatively
fixed set of features across timepoints, exhibiting less temporal
variation in feature selection. 
DIME concentrates acquisitions at later visits, consistent with its greedy selection criterion, which may undervalue early measurements.

\subsection{Robustness to Acquisition-Decision Perturbations and Distance from Learned Regions} 
\label{appendex:robustness}
We evaluate \OurMethod's robustness through two complementary analyses: (i) recovery from a deliberately perturbed acquisition decision and (ii) performance on test instances farther from the learned centroids.\looseness-1

\subsubsection{Robustness to Acquisition-Decision Perturbations}


\OurMethod revisits its acquisition recommendations as new observations become available. We assess whether this process remains effective after an early acquisition deviates from the policy's recommendation. At $t=0$, \OurMethod follows its standard policy. At $t=1$, we replace the recommended plan with a randomly selected cached plan from a different partition. From $t\geq 2$, \OurMethod resumes standard embedding-based planning using the resulting partial observations.

Table~\ref{tab:plan_perturbation} shows small changes in both AUROC and average acquisition cost across WOMAC, KLG, and ILIADD, suggesting that the overall performance--cost tradeoff is relatively insensitive to this single early perturbation.
This stability further illustrates \OurMethod's ability to adapt subsequent acquisition plans to the observations actually obtained.


\begin{table}[!ht]
\centering
\caption{WOMAC/KLG/ILIADD AUROC with average cost, without and with random-plan perturbation.}
\label{tab:plan_perturbation}
\small
\begin{tabular}{lrrrrrr}
\toprule
Dataset & \makecell{W/o perturb.\\AUROC} & \makecell{W/ perturb.\\AUROC} & $\Delta$ AUROC & \makecell{W/o perturb.\\cost} & \makecell{W/ perturb.\\cost} & $\Delta$ Cost \\
\midrule
KLG       & 0.8164 & 0.8101 & -0.0063 &  9.390 &  9.060 & -0.330 \\
WOMAC     & 0.6207 & 0.6246 &  0.0039 & 11.447 & 11.251 & -0.196 \\
ILIADD    & 0.8269 & 0.8258 & -0.0011 & 65.126 & 65.010 & -0.116 \\
\bottomrule
\end{tabular}
\end{table}



\subsubsection{Robustness Across Nearest-Centroid Distances}
\OurMethod retrieves cached acquisition plans by mapping each test state $(x_o,o,t)$ to its nearest centroid $\mu_{k,t}$ in the learned embedding space. We assess whether its performance remains stable for test instances farther from these centroids. For each instance, we compute the average nearest-centroid distance across all acquisition steps and compare the closest 80\% (\emph{inliers}) with the farthest 20\% (\emph{outliers}).

Table~\ref{tab:centroid_distance} shows small AUROC differences between the nearer and farther groups across WOMAC, KLG, and ILIADD, with relatively stable average acquisition costs. This stability suggests that cached acquisition plans remain useful even for test instances less closely matched to the learned centroids. Together with the plan-perturbation results, these findings support \OurMethod's use of a shared plan cache that accommodates variation in both acquisition histories and the partial observations encountered during inference.

\begin{table}[!ht]
\centering
\caption{WOMAC/KLG/ILIADD AUROC with average cost for the nearest and farthest nearest-centroid-distance groups.}
\label{tab:centroid_distance}
\small
\begin{tabular}{lrrrrrr}
\toprule
Dataset & \makecell{Inliers\\AUROC} & \makecell{Outliers\\AUROC} & $\Delta$ AUROC & \makecell{Inliers\\cost} & \makecell{Outliers\\cost} & $\Delta$ Cost \\
\midrule
KLG    & 0.8152 & 0.8188 &  0.0037 &  9.522 &  8.861 & -0.661 \\
WOMAC  & 0.6151 & 0.6059 & -0.0091 & 11.578 & 10.922 & -0.656 \\
ILIADD & 0.8309 & 0.8145 & -0.0164 & 65.617 & 63.160 & -2.456 \\
\bottomrule
\end{tabular}
\end{table}

%% file: sections/6_conclusion.tex

\section{Conclusion}
\label{sec:conclusion}
We introduced \texttt{\OurMethod}, a longitudinal AFA framework built on two complementary innovations: contrastive embeddings supervised by future acquisition-utility profiles, and a partitioning acquisition policy that caches plans over the learned space. To our knowledge, this is the first AFA framework to combine these mechanisms. NOCT guides both representation learning and offline plan selection, enabling states with similar acquisition needs to share planning effort. At deployment, nearest-centroid lookup supports fast acquisition decisions that adapt as new observations arrive. Across synthetic and real-world healthcare datasets, \OurMethod improves performance--cost tradeoffs over diverse LAFA baselines while enabling low-latency inference. These results highlight acquisition utility as a useful organizing principle for representation learning, connecting reusable nonmyopic planning with adaptive decisions about what to measure and when.

\section*{Acknowledgments}
This research was partly funded by National Science Foundation (NSF) grant DMS2324394 and by National Institutes of Health (NIH) grants 1R01AA02687901A1, 1OT2OD032581-02-321, and 1OT2OD038045-01. The views and conclusions contained in this document are those of the authors and should not be interpreted as representing official policies, either expressed or implied, of the NIH and NSF.

%% file: sections/7_appendix.tex



\newpage
\appendix
\onecolumn
\section*{Appendices and Supplemental Materials}

\section{Additional Discussion} \label{appendex:discussion}


\subsection{Broader impacts}
\label{appendex:impact}
In this paper, we introduce a nonmyopic active feature acquisition framework, \texttt{\OurMethod}, which could aid the decision-making process in real-world applications such as healthcare. Specifically, \texttt{\OurMethod} could provide a more efficient and cost-effective diagnosis by selectively acquiring essential medical examinations, thus minimizing patient burden (e.g., time and costs) through personalized and timely intervention. However, as in any ML system, using \texttt{\OurMethod} should involve thorough/robust evaluations and supervision by medical professionals to prevent any potential negative impact.

Moreover, we acknowledge that our experiments focus on the healthcare domain, an important application area; however, our method is not specific to any domain. Similar resource-constrained scenarios also exist in other domains, such as in energy-aware audio recognition \citep{monjur2023soundsieve} or IoT devices \citep{dunkels2011powertrace}. In fact, our AFA formulation has applications across a myriad of domains, including robotics (acquisition of sensors with resource constraints), education (acquisition of different exam types without overburdening students), and beyond.

\subsection{Assumptions \& Limitations}
\label{appendex:Limitations}
Like traditional AFA methods and RL-based baselines, \texttt{\OurMethod} assumes that the problem has a finite horizon with a limited number of time steps. Additionally, we acknowledge that, similar to other baselines presented in this paper, the performance of \texttt{\OurMethod} is sensitive to the data quality. In particular, poor label annotation in real-world clinical datasets can degrade the accuracy. However, extending to an infinite-horizon setting and mitigating the sensitivity to the data quality are beyond the scope of this paper. We will leave these challenges for future work.

\subsection{Time Complexity}
\label{appendex:Time}
With a total of $L$ time steps and a total number of features $M$, the time complexity is bounded by \uline{$O(ML^2)$ for RL baselines (RAS, AS, ASAC}), and \uline{$O(M^2L^2)$ for DiFA and DIME}. For ACO, the nearest-neighbor oracle baseline, deployment requires retrieving similar training instances and scoring candidate plans. A naive nearest-neighbor search over $n$ training instances costs $O(NML)$ per replanning step. After retrieval, scoring $|\mathcal{P}|$ candidate plans over the retrieved neighborhood $\mathcal{I}(x_o)$ gives a total trajectory complexity of \uline{$O(NML^2 + |\mathcal{I}(x_o)|ML^2|\mathcal{P}|)$ for the ACO} baseline, assuming $O(ML)$ computation per candidate plan per neighbor. During deployment, CAPE computes the current embedding, finds the nearest cached region, and retrieves the corresponding cached plan. With $K$ cached regions per timepoint and embedding dimension $E$, each nearest-centroid lookup costs $O(KE)$, and the full lookup overhead is at most \uline{$O(LKE)$ for \OurMethod}.

\subsection{Candidate-plan Construction $|\mathcal{P}|$}
\OurMethod does not enumerate the combinatorial plan space; instead, it evaluates a fixed sampled candidate set offline. For each partition $C_{k,t}$, we construct a candidate-plan set $\mathcal{P}_{k,t}$ whose elements are plans $v\subseteq\mathcal{V}_{>t}$, where $\mathcal{V}_{>t}$ contains all feature-time pairs after time $t$ and $|\mathcal{V}_{>t}| = M(L-t)$. For each candidate plan, we first sample its cardinality $q$ uniformly from the integers between $0$ and $M(L-t)$, and then uniformly sample $q$ distinct feature-time pairs from $\mathcal{V}_{>t}$. Repeating this procedure yields $|\mathcal{P}|$ binary acquisition masks $v\in\{0,1\}^{M\times(L-t)}$.

\subsection{Details on Acquisition-Utility Profile Construction $\zeta_{i,o}$}
\label{appendex:future_dist}
In this section, we formally introduce the acquisition-utility profiles $\zeta_{i,o}$. For any partial state $(x_{i,o},o,t)$ and corresponding labels $y_i$ from the training dataset $\mathcal{D}$, we identify the top-$r$ candidate plans $\hat v_{i,o}^{(1)}, \ldots, \hat v_{i,o}^{(r)} \subseteq \mathcal{V}_{>t}
= \{(m, t') \mid m \in \{1, \ldots, M\}, \, t' \in \{t+1, \ldots, L\}\}$, in ascending order of the plug-in objective $\widehat{\mathrm{NOCT}}_{\mathrm{pl}}(x_{i,o},o,\hat v_{i,o}^{(l)})$ (Eq.~(\ref{eq:plugobj})).
$\widehat{\mathrm{NOCT}}_{\mathrm{pl}}(x_{i,o},o,\hat v_{i,o}^{(l)})$ quantifies the accuracy--cost tradeoff for candidate plan $\hat v_{i,o}^{(l)}$, and we have access to $x_{i,\hat v_{i,o}^{(l)}}$ during training since it is from the training set. Each candidate plan $\hat v_{i,o}^{(l)}$ induces a uniform distribution $\nu_{\hat v_{i,o}^{(l)}}$ over its selected feature-time pairs (and termination), defined as $\nu_{\hat v_{i,o}^{(l)}}(m,t') = \tfrac{1}{|\hat v_{i,o}^{(l)}|}$ if $(m,t') \in \hat v_{i,o}^{(l)}$ and $0$ otherwise. Intuitively, $\nu_{\hat v_{i,o}^{(l)}}$ spreads equal probability mass over all acquisitions in the future candidate plan $\hat v_{i,o}^{(l)}$. When $\hat v_{i,o}^{(l)}$ contains no further acquisition, $\nu_{\hat v_{i,o}^{(l)}}(\varnothing) = 1$; otherwise, $\nu_{\hat v_{i,o}^{(l)}}(\varnothing) = 0$. The acquisition-utility profile $\zeta_{i,o}$ is the weighted sum of these distributions:
\begin{equation}
\zeta_{i,o} = \sum_{l=1}^{r} w_{i,o}^{(l)}\,\nu_{\hat v_{i,o}^{(l)}}
\label{eq:prob}
\end{equation}

where
\begin{equation*}
w_{i,o}^{(l)} =
\frac{\exp\!\big(-\widehat{\mathrm{NOCT}}_{\mathrm{pl}}(x_{i,o},o,\hat v_{i,o}^{(l)})\big)}
{\sum_{j=1}^{r}\exp\!\big(-\widehat{\mathrm{NOCT}}_{\mathrm{pl}}(x_{i,o},o,\hat v_{i,o}^{(j)})\big)}.
\end{equation*}
Intuitively, given a partial state $(x_{i,o},o,t)$, $\zeta_{i,o}$ indicates how likely each feature at each future timepoint (and termination) is to be selected by weighting its accuracy--cost tradeoff.

\subsection{Proof of Theorem 3.1: $-\mathrm{NOCT}$ Lower Bounds the Optimal Longitudinal AFA MDP Value}
\label{appendex:theorem}

\textbf{Longitudinal AFA modeled as an MDP}\quad Longitudinal AFA can be formulated as a Markov Decision Process (MDP). The state $\mathbf{s}_t = \bigl({x}_{o},\, o,\,  t\bigr)$, where ${x}_{o}$ denotes the (partial) observations through time $t$ and $o \subseteq \{(m, t') \mid m \in \{1, \ldots, M\}, \, t' \in \{1, \ldots, t\}\}$ records which features were acquired.
We suppose access to a pre-trained arbitrary conditioning classifier $\hat{y}$ ~\citep{shim2018joint, li2021active} that can predict the label at any target timepoint $k$ with the current information, i.e., $\hat{y}_k(x_o)$. Note that the same input information can be used to predict at other time points, $\hat{y}_{k^\prime}(x_o)$ (where forward prediction corresponds to when feature observations in $o$ occur before $k^\prime$).
The action space is $a \in \{a_{t'}\}_{t'=t+1}^L \cup \{\varnothing\}$, where $a_{t'} \in \{0,1\}^M$ indicates which of the $M$ features to acquire at time $t'$, or, when $a = \varnothing$, to terminate and predict all remaining timepoints. For a non-terminal action, the transition is $\bigl({x}_{o},\, o, \, t\bigr) \rightarrow \bigl({x}_{o \cup a_{t'}},\, o \cup a_{t'},\, t' \bigr)$, where (for notational simplicity) $o \cup a_{t'}$ denotes the union of the previous acquisitions and tuples of features at time $t'$ in $a_{t'}$.
Since this implies that we are not acquiring any information until time $t'$, the reward is the negative cross-entropy (CE) loss of making predictions up to $t'-1$ (i.e., $\hat{y}_{t+1}, \ldots, \hat{y}_{t'-1}$) with previous information (${x}_{o}$) and the respective negative CE loss and cost of acquiring new features at time $t'$:
$-\sum_{k=t+1}^{t'-1} \mathrm{CE}(\hat{y}_k({x}_{o}),\,y_{k}\bigr) - \mathrm{CE}(\hat{y}_{t'}({x}_{o \cup a_{t'}}),\,y_{t'}\bigr) -\alpha\sum_{m=1}^M a_{t',m}c^m$,
where $\alpha$ is an \emph{application-specific} accuracy-cost tradeoff parameter; certain applications allow a higher cost (i.e., lower $\alpha$) and some are more cost-restricted (i.e., high $\alpha$). When terminated ($a = \varnothing$) at timepoint $t$, we are not acquiring any additional information for the remaining timepoints; thus, the reward is $-\sum_{k=t+1}^L \mathrm{CE}(\hat{y}_k({x}_{o}),\,y_{k}\bigr)$, which represents the negative cross-entropy loss accumulated from $t+1$ to $L$ using the current information at hand.
We denote our policy as $\pi\bigl({x}_{o},\, o,\, t\bigr)$, which selects
actions to maximize the expected cumulative rewards through $L$.

\textbf{Theorem 3.1.}
Let $V^\star(\mathbf{s})$ be the optimal value of the corresponding finite-horizon LAFA. Consider a partially observed state $\mathbf{s}=(x_o,o,t)$ given a fixed predictor $\hat{y}$ and nonnegative acquisition costs. We have:
\begin{equation}
\text{If}\ J_{\mathrm{NOCT}}(\mathbf{s})
=
-\min_{v\subseteq \mathcal{V}_{>t}} \mathrm{NOCT}(x_o,o,v),\  \text{then}\  J_{\mathrm{NOCT}}(\mathbf{s})\le V^\star(\mathbf{s}).
\end{equation}

\uline{Problem Setup.} We extend the argument of \citet[Theorem~C.1]{valancius2024acquisition} to the longitudinal setting. For any observations $(x_o, o, t)$, let the set of remaining feature-time pairs that are still available after time $t$ be:
\begin{equation*}
\mathcal{V}_{>t} = \{(m, t') \mid m \in \{1, \ldots, M\}, \, t' \in \{t+1, \ldots, L\}\}.
\end{equation*}
To follow the convention of reinforcement learning literature, we redefine the objective NOCT (Eq.~(\ref{eq:obj})) as maximization instead of minimization without loss of generality. For notational convenience, we write $\texttt{NOCT-MAX}^s_t(x_o)\equiv \texttt{NOCT-MAX}^s_t(x_o,o)$, since $x_o$ is indexed by the observation set $o$. That is, for any partial observation $(x_o,o,t)$, we define $\texttt{NOCT-MAX}^s_t(x_o)$ as:

\begin{equation*}
\texttt{NOCT-MAX}_t^s(x_o) := \max_{\substack{v\subseteq \mathcal{V}_{>t} \\ \text{s.t } \mid\text{rounds}(v) \mid \leq s} }
\Bigl[ - \mathbb{E}_{y_{t+1:L}, x_v \mid {x}_{o} }[\ell (x_o \cup x_v, y, t)] - \alpha \!\!\!\! \sum_{(m, t') \in v} \!\!\!\! c^m \Bigr],
\end{equation*} 
where $s \in \mathbb{N}$ represents the maximum number of possible future acquisition rounds, and let $\text{rounds}(v) := \{ t' \mid  \exists m \in \{1,\ldots,M\}:(m, t') \in v \}$. We then define 
\begin{equation*}
V_t^0(x_o) := -\mathbb{E}_{y_{t+1:L}\mid x_o}[\ell(x_o, y, t)],
\end{equation*}
\[
\begin{aligned}
\Delta_{t,t'}(x_o,a_{t'})
:={}&\mathbb E_{y_{t+1:t'},\,x_{a_{t'}}\mid x_o}
\Bigg[
\sum_{k=t+1}^{t'-1}\mathrm{CE}\bigl(\hat y_k(x_o),y_k\bigr) +\mathrm{CE}\bigl(\hat y_{t'}(x_{o\cup a_{t'}}),y_{t'}\bigr)
\Bigg],
\end{aligned}
\]
and 

\begin{align}
V_t^s(x_o) &:= \max \Bigl\{V_t^0(x_o), \max_{\substack{t' \in \{t+1, ..., L\} \\ a_{t'} \in \{0,1\}^M}} \Bigl[-\Delta_{t,t'}(x_o,a_{t'}) \nonumber
\\ &\qquad -\alpha \sum_m a_{t',m}c^m + \mathbb{E}_{x_{a_{t'}}\mid x_o} [V^{s-1}_{t'}(x_{o \cup a_{t'}})] \Bigr] \Bigr\}.
\label{eq:value_func}
\end{align}

Here, $V_t^s(x_o)$ denotes the optimal expected cumulative reward from time $t+1$ through $L$, given the current partial observations $x_o$, with at most $s$ further acquisition rounds.
We see that $V^{0}_{t}$ is the expected negative cumulative loss when we immediately terminate at $t$, and we predict the future labels using only what we have acquired so far, $x_o$. Thus, $V^{s}_{t}$ is the value for the optimal longitudinal AFA policy, such that it either (i) terminates, returning $V_t^0$ or (ii) executes another acquisition at the future time step $t' \in (t+1,...,L)$ and acquires the feature $a_{t'}\in \{0,1\}^M$. 

\uline{Claim.}
For every $(x_o, o, t)$ with $s \in \mathbb{N}_{\geq 0}$, we have $V_t^s(x_o) \geq \texttt{NOCT-MAX}^s_t(x_o)$.

\uline{Proof.}

\emph{Base case $s=0$.} Since there is no further action, both the MDP agent and \texttt{NOCT-MAX} terminate:
\begin{equation*}
V_{t}^0(x_o) =  -\mathbb{E}_{y_{t+1:L}\mid x_o}[\ell (x_o, y, t)] = \texttt{NOCT-MAX}^0_t(x_o).
\end{equation*}
\emph{Inductive hypothesis.} Assume for $s-1 \geq 0$, we have: 
\begin{equation}
V_{t'}^{s-1}(x) \geq \texttt{NOCT-MAX}^{s-1}_{t'}(x) \quad \forall (x,t').
\tag{IH}
\end{equation}
\emph{Inductive step.} We fix $(x_o, t)$ and see that: 
\begin{align*}
&V_t^s(x_o)
\\ &= \max \Bigl\{V_t^0(x_o), \max_{\substack{t' \in \{t+1, ..., L\} \\ a_{t'} \in \{0,1\}^M}} \Bigl[ -\Delta_{t,t'}(x_o,a_{t'}) -\alpha \sum_{m} a_{t',m}c^m
\\ &\qquad + \mathbb{E}_{x_{a_{t'}}\mid x_o} [V^{s-1}_{t'} (x_{o \cup a_{t'}} )]  \Bigr] \Bigr\}
\\
& \!\!\overset{(\text{IH})}{\geq} \max \Bigl\{V_t^0(x_o), \max_{\substack{t' \in \{t+1, ..., L\} \\ a_{t'} \in \{0,1\}^M}} \Bigl[ -\Delta_{t,t'}(x_o,a_{t'}) -\alpha \sum_{m} a_{t',m}c^m
\\ &\qquad + \mathbb{E}_{x_{a_{t'}}\mid x_o} [\texttt{NOCT-MAX}^{s-1}_{t'} (x_{o \cup a_{t'}} )]  \Bigr] \Bigr\}
\\
&= \max \Bigl\{V_t^0(x_o), \\ &\max_{\substack{t' \in \{t+1, ..., L\} \\ a_{t'} \in \{0,1\}^M}} \Bigl[-\Delta_{t,t'}(x_o,a_{t'}) -\alpha \sum_{m} a_{t',m}c^m + \mathbb{E}_{x_{a_{t'}}\mid x_o} \Bigl[\!\!\!\!\max_{\substack{v'\subseteq \mathcal{V}_{>t'} \\ \text{s.t } \mid\text{rounds}(v') \mid \leq s-1} }
\!\!\!\!\!\!\!\!\!\!\!\Bigl(
\\ &\qquad - \mathbb{E}_{y_{t'+1:L}, x_{v'} \mid {x}_{o}, x_{a_{t'}} }[\ell (x_o \cup x_{a_{t'}}\cup x_{v'}, y, t')] - \alpha \!\!\!\!\!\! \sum_{(m, t'') \in v'} \!\!\!\!\!\!c^m \Bigr)\Bigr]  \Bigr] \Bigr\}
\\
&\overset{*}\geq \max \Bigl\{V_t^0(x_o), \\
&\max_{\substack{t' \in \{t+1, ..., L\} \\ a_{t'} \in \{0,1\}^M}} \max_{\substack{v'\subseteq \mathcal{V}_{>t'} \\ \text{s.t } \mid\text{rounds}(v') \mid \leq s-1} } \Bigl[
\\ &\qquad - \mathbb{E}_{y_{t'+1:L}, x_{v'}, x_{a_{t'}}\mid {x}_{o} }[\ell (x_o \cup x_{a_{t'}}\cup x_{v'}, y, t')]
\\ &\qquad - \Delta_{t,t'}(x_o,a_{t'}) - \alpha \Bigl(\sum_{(m, t'') \in v'} \!\!\!\! c^m + \sum_{m} a_{t',m}c^m \Bigr) \Bigr] \Bigr\}
\\
&= \max\Bigl\{V_t^0(x_o), \max_{\substack{v \subseteq \mathcal{V}_{>t} \\ \text{s.t } |\text{rounds}(v)| \le s} }
[ - \mathbb{E}_{y_{t+1:L}, x_v \mid {x}_{o} }[\ell (x_o \cup x_v, y, t)] - \alpha \!\!\!\! \sum_{(m, t'') \in v} \!\!\! c^m]\Bigr\}
\\
&= \texttt{NOCT-MAX}^{s}_{t} (x_o).
\end{align*}
where * follows from:
\[
\begin{aligned}
&\begin{array}{r}
\forall v' \in \Omega, x_{a_{t'}},\  \max\limits_{v \in \Omega} \mathcal{L}\left(x_{a_{t'}}, v\right) \geq \mathcal{L}\left(x_{a_{t'}}, v'\right) \quad \Longrightarrow \\[0.8em]
\forall v'\in \Omega,\ \mathbb{E}_{x_{a_{t'}} \mid x_o}\left[\max\limits _{v \in \Omega} \mathcal{L}\left(x_{a_{t'}}, v\right)\right] \geq \mathbb{E}_{x_{a_{t'}} \mid x_o}\left[\mathcal{L}\left(x_{a_{t'}}, v'\right)\right] \Longrightarrow \\[0.8em]
\mathbb{E}_{x_{a_{t'}} \mid x_o}\left[\max\limits _{v \in \Omega} \mathcal{L}\left(x_{a_{t'}}, v\right)\right] \geq \max\limits _{v \in \Omega} \mathbb{E}_{x_{a_{t'}} \mid x_o}\left[\mathcal{L}\left(x_{a_{t'}}, v\right)\right]
\end{array}
\end{aligned}
\]
for $\Omega :=\Bigl\{v\subseteq\mathcal V_{>t'}\;\bigm|\; |\text{rounds}(v)|\le s-1\Bigr\}$ and 
\[
\mathcal{L}(x_{a_{t'}},v) :=
-\mathbb{E}_{y_{t'+1:L},x_v\mid x_o,x_{a_{t'}}}
[\ell(x_o \cup x_{a_{t'}} \cup x_v,y,t')]
-\alpha \left(
\sum_{(m,t'')\in v} c^m + \sum_m a_{t',m}c^m
\right).
\]

Note that one may recover the non-acquisition-round constrained problem by considering large enough $s$, yielding a lower bound on the longitudinal AFA MDP.

Theorem~3.1 establishes why NOCT is an appropriate objective for evaluating future acquisition plans. Under a fixed predictor and nonnegative acquisition costs, the negative of the minimum ideal expected NOCT objective provides a lower bound on the optimal finite-horizon LAFA value. Thus, NOCT is not an ad hoc acquisition score: it is a principled plan-level surrogate tied to the same horizon-wide prediction-loss and acquisition-cost tradeoff as the sequential LAFA problem.

\subsection{Longitudinal AFA MDP Challenges}
Modeling longitudinal AFA as an MDP introduces several difficulties: (1) \emph{a large and complicated action space}, since one requires decisions on when to acquire and which features to acquire; (2) \emph{a complicated state space}, as it evolves dynamically over the combinatorial acquisitions as new features are captured over time; and (3) \emph{a credit assignment problem}, since acquisitions are rewarded at an aggregate level, making it hard to disentangle exactly which of the acquired features were most responsible for prediction rewards \citep{li2021active}. Despite these challenges, prior work \citep{qin2024risk, yoon2019asac, kossen2022active, nguyen2024active} often relies on RL for longitudinal AFA. In contrast, \texttt{\OurMethod} is grounded in an MDP formulation but does not learn a Bellman value function, use temporal-difference learning, or optimize an actor-critic policy. Instead, we define NOCT as an explicit plan-level objective over future acquisition trajectories. The ideal NOCT objective induces a lower bound on the optimal longitudinal AFA MDP value, and \OurMethod uses plug-in NOCT scores from training trajectories to learn an acquisition-utility manifold and fit a partitioning acquisition policy.

\section{Additional results}
\label{sec:addl_results}

Please find additional results below.

\begin{figure*}[t] 
  \centering
  \includegraphics[width=\textwidth]{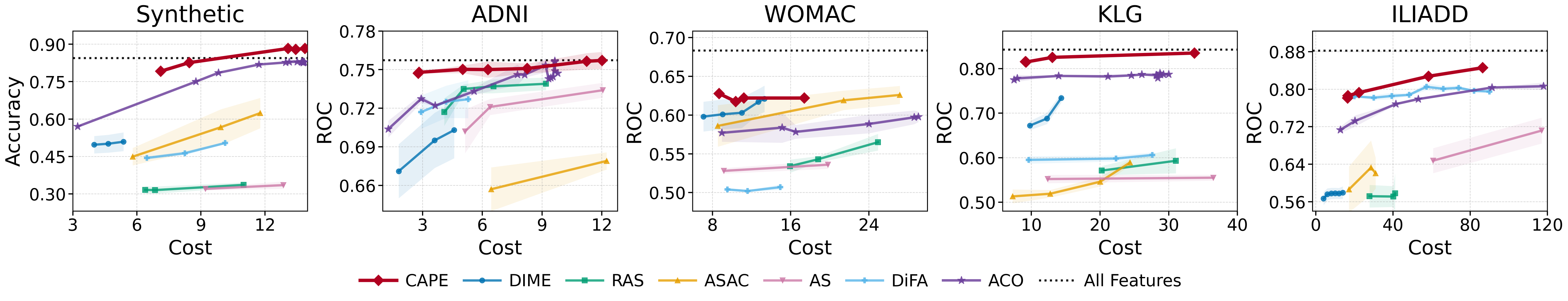}
  \vspace{-1.7em}
  \caption{Performance/cost of models across various average acquisition costs (budgets), with standard error across 5 runs. Following \citet{kossen2022active}, we report accuracy instead of AUROC for synthetic data. The dotted horizontal line shows performance using all features. \OurMethod consistently yields higher predictive performance at comparable or lower acquisition costs.}
  \label{fig:all_performance}
  \vspace{-8pt}
\end{figure*}
\begin{figure*}[t]
  \centering
      \includegraphics[trim={0cm 9.6cm 7.55cm 0cm}, clip, width=0.9\textwidth]{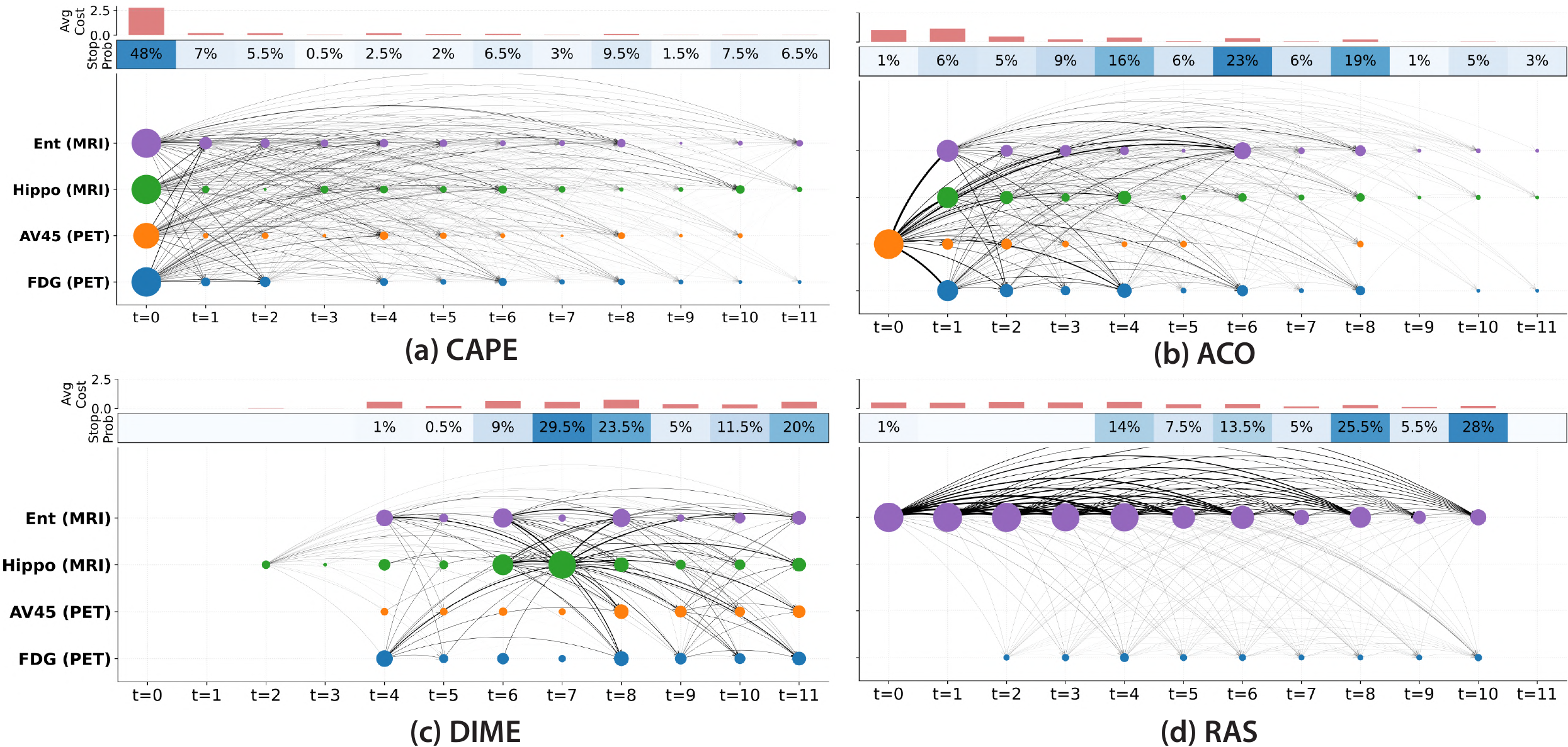}
      \vspace{-0.5em}
  \caption{Acquisition dynamics on the test set of the ADNI dataset. For each method, \textbf{top} shows the average acquisition cost at each timepoint (mean per sample), and \textbf{middle} shows the termination distribution that occurred at each timepoint. The \textbf{bottom} panel visualizes feature-time acquisition as a directed graph: each node corresponds to a feature acquired at a specific timepoint, and node size shows how often that feature is acquired at that time. A directed arrow from $(m,t)$ to $(m',t')$ (with $t'>t$) indicates the frequent co-acquisition across time, i.e., samples that acquire $(m,t)$ also acquire $(m',t')$ later. Average acquisition cost per sample: $4.0375$ (\OurMethod), $3.985$ (ACO), $4.0475$ (DIME), and $3.9750$ (RAS). \looseness-1}
  \label{fig:adni_qual_dynamics}
\end{figure*}
\begin{figure*}[t]
    \centering
    \vspace{0pt}
    \includegraphics[trim=0pt 0pt 0cm 0pt, clip, width=.7\linewidth]{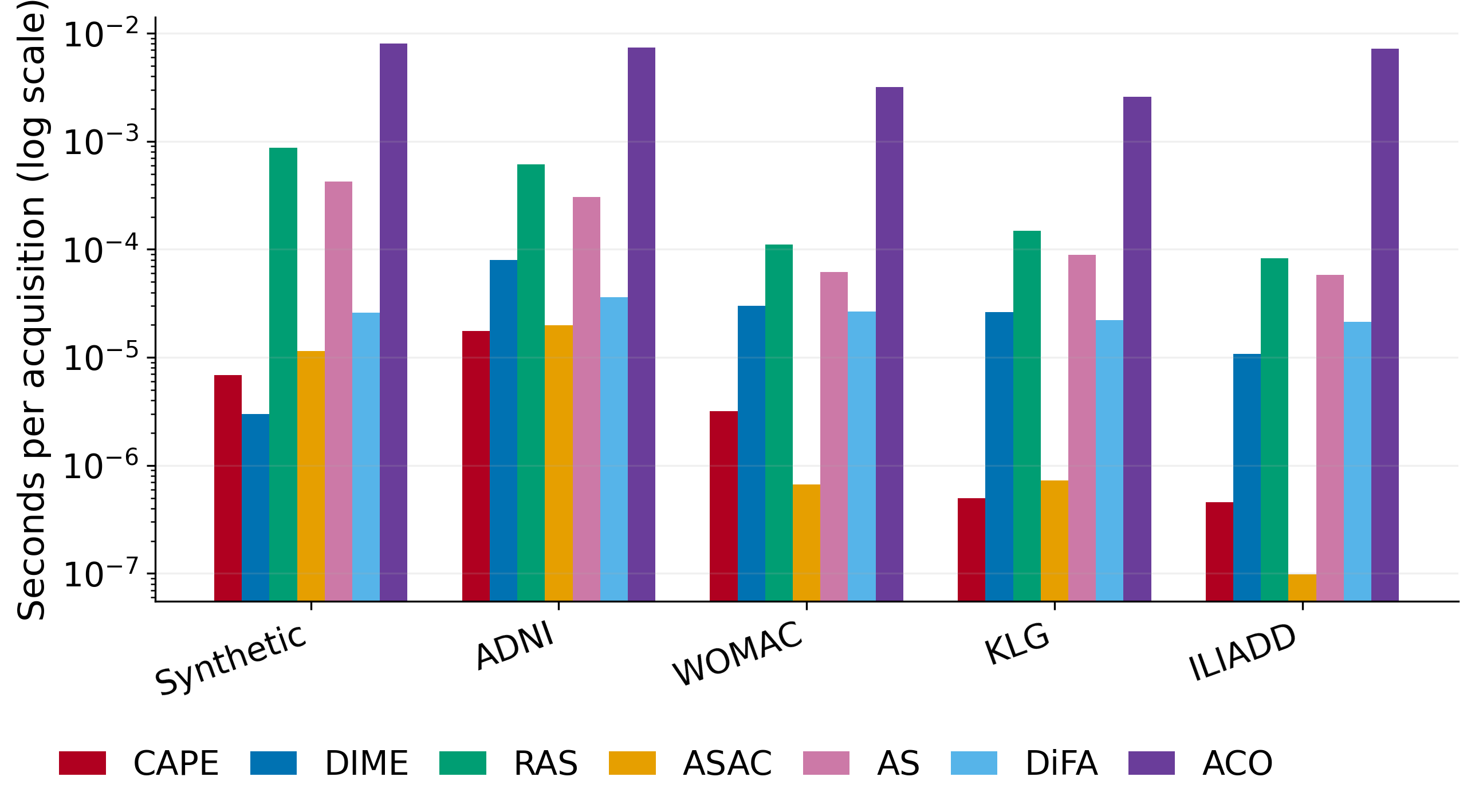}
    \vspace{0.0em}
    \caption{Inference latency per acquisition.}
    \label{fig:all_timing_eval}
    \vspace{-10pt}
\end{figure*}

\textbf{ADNI}\quad We additionally consider the Alzheimer's Disease Neuroimaging Initiative (ADNI) dataset (\url{https://adni.loni.usc.edu})~\citep{petersen2010alzheimer}, a longitudinal, multi-center observational study of cognitive decline. Following~\citet{qin2024risk}, we use $N=1{,}002$ patients diagnosed as cognitively unimpaired, mild cognitive impairment, or Alzheimer's disease~\citep{o2008staging,o2010validation}, with four PET/MRI biomarkers (FDG, AV45, Hippocampus, and Entorhinal) across $L=12$ visits. We split the dataset into train/val/test ($64\%/16\%/20\%$).
\looseness-1
Each patient may have a different pattern of missing features or labels, and we only report the results based on those that have a diagnosis label. 
We observe similar patterns, e.g., Figs.~\ref{fig:all_performance}--\ref{fig:all_timing_eval}.

\begin{table*}[t]
\captionsetup{skip=2pt}
\caption{Ablation comparing acquisition-aware embeddings $g_\phi(\cdot)$ used by \texttt{CAPE} against partitioning directly in the raw feature-value space. We report the mean $\pm$ standard errors across five runs.}
\label{tab:appx_embedding_ablation}
\begin{center}
\begin{sc}
\resizebox{1\linewidth}{!}{%
\begin{tabular}{c
                cc @{\quad}
                cc @{\quad}
                cc @{\quad}
                cc @{\quad}
                cc}
\toprule
\multirow{2}{*}{Method}
  & \multicolumn{2}{c}{Synthetic}
  & \multicolumn{2}{c}{ADNI}
  & \multicolumn{2}{c}{WOMAC}
  & \multicolumn{2}{c}{KLG}
  & \multicolumn{2}{c}{ILIADD} \\
\cmidrule(lr){2-3} \cmidrule(lr){4-5} \cmidrule(lr){6-7} \cmidrule(lr){8-9} \cmidrule(lr){10-11}
 & Acc. & cost
 & AUROC & cost
 & AUROC & cost
 & AUROC & cost
 & AUROC & cost \\
\midrule
\multirow{2}{*}{\makecell{Feature\\Partitioning}}
 & $0.774\pm0.000$ & $7.178\pm0.005$
 & $0.739\pm0.000$ & $2.760\pm0.011$
 & $0.618\pm0.005$ & $8.461\pm0.055$
 & $0.804\pm0.002$ & $9.912\pm0.046$
 & $0.762\pm0.009$ & $17.283\pm0.985$ \\
 & $0.866\pm0.000$ & $13.932\pm0.004$
 & $0.746\pm0.002$ & $12.328\pm0.107$
 & $0.610\pm0.004$ & $17.073\pm0.204$
 & $0.823\pm0.001$ & $34.931\pm0.053$
 & $0.822\pm0.004$ & $81.634\pm1.818$ \\
\midrule
\multirow{2}{*}{\texttt{CAPE}}
 & $\mathbf{0.792}\pm\mathbf{0.000}$ & $7.114\pm0.005$
 & $\mathbf{0.748}\pm\mathbf{0.001}$ & $2.821\pm0.020$
 & $\mathbf{0.628}\pm\mathbf{0.004}$ & $8.673\pm0.275$
 & $\mathbf{0.815}\pm\mathbf{0.004}$ & $9.147\pm0.082$
 & $\mathbf{0.786}\pm\mathbf{0.001}$ & $16.672\pm1.717$ \\
 & $\mathbf{0.883}\pm\mathbf{0.000}$ & $13.877\pm0.008$
 & $\mathbf{0.757}\pm\mathbf{0.007}$ & $12.012\pm0.208$
 & $\mathbf{0.622}\pm\mathbf{0.004}$ & $17.395\pm0.204$
 & $\mathbf{0.835}\pm\mathbf{0.003}$ & $33.758\pm1.271$
 & $\mathbf{0.846}\pm\mathbf{0.005}$ & $86.450\pm1.818$ \\
\bottomrule
\end{tabular}%
}
\end{sc}
\end{center}
\end{table*}

\input{sections/7_additional_results}

\section{Experimental Setup} \label{appendex:experiment_setup}
\subsection{Cost of Features} \label{sec:cost}
\textbf{Synthetic data.} Our synthetic dataset includes $2$ features, one being the digit and the other being the counter. Since both are generated through random selection, we assign an equal cost, $1$, to the two features. Upon publication, we will release the synthetic dataset. 

\textbf{ADNI.} The ADNI dataset includes $4$ features per time point, with FDG and AV45 extracted from PET scans and Hippocampus and Entorhinal extracted from MRI scans. Since PET is a more expensive diagnosis, we follow~\citet{qin2024risk} and assign a cost of $1$ to FDG and AV45, and a cost of $0.5$ to Hippocampus and Entorhinal. Access to ADNI data may be requested via \url{https://adni.loni.usc.edu}, and the Data Use Agreement is available at \url{https://adni.loni.usc.edu/terms-of-use/}.

\textbf{OAI.} We use a total of $27$ features for the OAI dataset, with $17$ clinical measurements and $10$ joint space width (JSW) measurements extracted from knee radiography. For the clinical measurements, we assign a low cost of $0.3$ to those that require minimal effort, e.g., age, sex, and race, and a slightly higher cost of $0.5$ to blood pressure and BMI calculation. For JSW, a cost of $1.0$ is assigned for the minimum JSW and $0.8$ for those measured at different positions. Access to the OAI data may be requested via \url{https://nda.nih.gov/oai/}.

\textbf{ILIADD.} We use a total of $34$ features for the ILIADD dataset \citep{Ringwald2025DoYF}, with features consisting of demographics and scales derived from the HiTOP-SR, factor scales, symptom scales, and alcohol use scales, and temporal features assessing momentary positive affect, energy, stress, and impulsivity. For all measurements, we assign an equal cost of $1.0$. 

\subsection{Baseline Implementation} \label{sec:baseline}
For ASAC, RAS, and AS, we use the implementation available at \url{https://github.com/yvchao/cvar_sensing}, which is under the BSD-3-Clause license. For DIME~\citep{gadgil2023estimating} and ACO~\citep{valancius2024acquisition}, we extend the methods to the longitudinal setting by building on top of the official implementation available at \url{https://github.com/suinleelab/DIME} and \url{https://github.com/lupalab/aaco}, respectively. At the time of writing, the DiFA codebase is not publicly released, and researchers may request it directly from the authors.

Following~\citet{qin2024risk}, we share the same neural CDE predictor ~\citep{kidger2020neural} for ASAC, RAS, and AS. Following the authors' released code, we treat the drop rate $p$ of the auxiliary observation strategy $\pi_0$ (the same notation as in ~\citet{qin2024risk}) as a hyperparameter and sweep $p \in \{0.0, 0.3, 0.5, 0.7\}$. We select $p$ based on the outcome predictor's average accuracy over five randomly masked validation sets generated by $\pi_0$. This matches the original evaluation protocol to ensure comparability and reproducibility. The selected drop rates are $p=0.3$ (Synthetic), $p=0.7$ (ADNI), $p=0.5$ (WOMAC), $p=0.3$ (KLG), and $p=0.7$ (ILIADD).

\textbf{ASAC~\citep{yoon2019asac}.} We select the coefficient for the acquisition cost: 1) $\mu \in \{0.0015, 0.002, 0.003\}$ for the synthetic dataset, 2) $\mu \in \{0.005, 0.02\}$ for ADNI, 3) $\mu \in \{0.0012, 0.0016, 0.003\}$ for the WOMAC of OAI dataset, 4) $\mu \in \{0.00125, 0.0015, 0.00175, 0.005\}$ for the KLG of OAI dataset, and 5) $\mu \in \{0.02, 0.005, 0.002\}$
for ILIADD. Note that we report multiple final $\mu$ values because we evaluate ASAC across multiple acquisition costs.

\textbf{RAS~\citep{qin2024risk}.} For the synthetic dataset, we set the allowed acquisition interval to $(\Delta_{\min} = 0.2, \Delta_{\max} = 1.0)$ for the synthetic dataset, $(\Delta_{\min} = 0.5, \Delta_{\max} = 1.5)$ for ADNI, and $(\Delta_{\min} = 0.5, \Delta_{\max} = 4.5)$ for WOMAC, KLG, and ILIADD. 

Moreover, for RAS, we tune the diagnostic-error coefficient $\gamma^{\mathrm{RAS}}$
on the validation set using a dataset-specific grid search. Table~\ref{tab:ras_gamma}
lists the search grids and the final selected values. We report multiple selected
values because RAS is evaluated across multiple acquisition-cost settings. For all
datasets, we use a tail-risk quantile of $0.1$, an invalid-visit penalty of $10$, and
a discount factor of $0.99$, following the authors' defaults.

\begin{table}[t]
\centering
\caption{RAS diagnostic-error coefficient grids and selected values.}
\label{tab:ras_gamma}
\small
\begin{tabular}{lll}
\toprule
Dataset & Search grid for $\gamma^{\mathrm{RAS}}$ & Selected $\gamma^{\mathrm{RAS}}$ \\
\midrule
Synthetic & $\{200,250,280,300,310,320,350,400,5000,7000,9000\}$ & $\{5000,7000,9000\}$ \\
ADNI & $\{50,75,100,175,200,250,300,350,400,450\}$ & $\{50,75,100,175\}$ \\
WOMAC & $\{200,250,300,350,400,450,2000,3000,4500,9000\}$ & $\{2000,4500,9000\}$ \\
KLG & $\{200,250,300,350,400,450,1200,1700\}$ & $\{1200,1700\}$ \\
ILIADD & $\{2,5,10,15,20,30\}$ & $\{5,10,15\}$ \\
\bottomrule
\end{tabular}
\end{table}

\textbf{AS~\citep{qin2024risk}.} We set the minimum and maximum allowed acquisition intervals as in the RAS setting. In addition, we select a fixed acquisition interval of: 1) $\tilde{\Delta} \in \{0.2, 0.4\}$ for the synthetic dataset, 2) $\tilde{\Delta} \in \{0.5, 1.0, 1.5\}$ for ADNI and ILIADD datasets, 3) $\tilde{\Delta} \in \{0.5, 1.5\}$ for WOMAC of OAI, and 4) $\tilde{\Delta} \in \{0.5, 1.5\}$ for KLG of OAI dataset. Note that we report multiple final $\tilde{\Delta}$ because we evaluate AS across multiple acquisition costs. 

\textbf{DIME~\citep{gadgil2023estimating}.} For this baseline, we extend the original model to the longitudinal setting, while still keeping its greedy feature acquisition strategy. 
Both the prediction network and value network of DIME share the same architecture as our \OurMethod prediction model, and hence we use the same training strategies. 

\textbf{DiFA~\citep{ghosh2023difa}}. Following ~\citet{ghosh2023difa}, we use a variational autoencoder with arbitrary conditioning (VAEAC) probabilistic model  \citep{ivanov2018variational} for imputation. When using the same architecture with \OurMethod, DiFA performs poorly. For a stronger DiFA baseline, we use the architecture recommended in the DiFA paper. That is, both the prediction network and the feature policy model have two fully connected layers (hidden size 128) with skip connections, dropout regularization, and LeakyReLU activation function. 

\textbf{ACO~\citep{valancius2024acquisition}.} We use the same prediction network and acquisition costs as \OurMethod (see Appx.~\ref{sec:architecture} for details). For the nearest neighbor search, we use $10$ neighbors across all datasets. To extend ACO to the longitudinal setting, the method selects the earliest available subset of features and then replans based on the acquired features.

\subsection{\texttt{\OurMethod} Implementation} \label{sec:architecture}
\textbf{Prediction Network.} We use an MLP architecture for \OurMethod's prediction network. Specifically, the network consists of two hidden layers with ReLU. Each hidden layer contains $10$ units for the synthetic and ADNI datasets. For WOMAC, KLG, and ILIADD, we use a hidden unit of $32$. Moreover, we apply random dropout to the input during training of the predictor.

For the embedding network, we use four hidden layers with ReLU. Each hidden layer contains $32$ units for the synthetic and ADNI datasets. For WOMAC, KLG, and ILIADD, we use a hidden unit of $64$.

We select the hyperparameter $\alpha$ in Eq.~(\ref{eq:obj}) to vary the performance-cost tradeoff. For the synthetic dataset, we use $\alpha \in \{0.0009, 0.002, 0.006, 0.008, 0.01, 0.02, 0.03\}$. For ADNI, we use $\alpha \in \{0.0007, 0.001, 0.003, 0.006, 0.01, 0.05, 0.07\}$. For WOMAC of OAI, we use $\alpha \in \{0.00075, 0.0025, 0.005, 0.006, 0.009\}$. For KLG of OAI, we use $\alpha \in \{0.00075, 0.001, 0.0025, 0.005, 0.006\}$. For ILIADD, we use $\alpha \in \{0.1, 0.07, 0.03, 0.01, 0.007, 0.003, 0.001, 0.0007, 0.0003, 0.0001, 7\times10^{-5}, 3\times10^{-5}, 10^{-5}, 3\times10^{-6}, 10^{-6}\}$.

\paragraph{Role of the budget parameter $\alpha$.}
$\alpha$ specifies the desired cost--benefit operating point; it is not a test-time threshold. As in many prior cost-sensitive AFA methods---including RAS, AS, ASAC, and ACO---a separate policy is trained for each cost--benefit tradeoff. Since $\alpha$ affects the top-$r$ acquisition-utility profiles $\zeta_{i,o}$, learned embedding $g_\phi$, and cached plans $\hat v_{k,t}^{*}$, our implementation trains a separate embedding and constructs a separate cache for each $\alpha$.

\OurMethod's ``adaptability'' refers to within-trajectory replanning, not adaptation to arbitrary cost preferences using a single cache. After each new acquisition step, \OurMethod re-embeds the updated partial state $(x_o,o,t)$ using $g_\phi$ and may retrieve a different cached plan $\hat v_{k,t}^{*}$ based on the realized trajectory, rather than committing to the complete plan selected previously. Accordingly, each point on the reported frontier corresponds to an $\alpha$-specific policy for both \OurMethod and the cost-sensitive baselines.


\subsection{Hardware and Timing} \label{sec:hardware}
All experiments were conducted using a single NVIDIA L40S GPU and an Intel Xeon Gold 6526Y CPU. To prevent inaccuracies from asynchronous CUDA execution, device synchronization was enforced before and after all timed regions, with durations recorded via \texttt{time.perf\_counter()}.

%% file: sections/7_additional_results.tex
\subsection{Hyperparameter Sensitivity}
\label{appendex:hyperparameter_sensitivity}
We evaluate sensitivity to the contrastive margin $\gamma$, divergence weight $\beta$, embedding dimension $E$, and the number of top plans $r$, while holding all other settings fixed. As shown in Tables~\ref{tab:sensitivity_gamma}--\ref{tab:sensitivity_r}, \OurMethod remains highly robust across a broad range of configurations, indicating that our method can be readily applied without excessive hyperparameter tuning. Furthermore, across these varied settings, \OurMethod consistently maintains its performance advantage over the baselines reported in Figure~\ref{fig:all_performance}.

\begin{table}[!ht]
\centering
\caption{Test performance and average cost by contrastive-loss margin $\gamma$ and cost regime.}
\label{tab:sensitivity_gamma}
\scriptsize
\setlength{\tabcolsep}{2.7pt}
\resizebox{\textwidth}{!}{%
\begin{tabular}{cc*{4}{cc}}
\toprule
$\gamma$ & Cost regime & ADNI AUROC & ADNI Cost & WOMAC AUROC & WOMAC Cost & KLG AUROC & KLG Cost & ILIADD AUROC & ILIADD Cost \\
\midrule
0.5 & Low  & 0.7468 & 2.788  & 0.6236 & 8.981  & 0.8197 & 9.359  & 0.7926 & 19.543 \\
1   & Low  & 0.7488 & 2.768  & 0.6201 & 8.980  & 0.8168 & 9.638  & 0.7931 & 19.780 \\
2   & Low  & 0.7507 & 2.772  & 0.6224 & 8.822  & 0.8188 & 9.984  & 0.7956 & 19.338 \\
\midrule
0.5 & High & 0.7539 & 11.295 & 0.6256 & 18.191 & 0.8363 & 38.084 & 0.8453 & 89.853 \\
1   & High & 0.7547 & 12.002 & 0.6241 & 17.245 & 0.8353 & 38.299 & 0.8454 & 90.817 \\
2   & High & 0.7548 & 12.105 & 0.6254 & 17.457 & 0.8351 & 37.550 & 0.8459 & 89.042 \\
\bottomrule
\end{tabular}}
\end{table}

\begin{table}[!ht]
\centering
\caption{Test performance and average cost by divergence weight $\beta$ and cost regime.}
\label{tab:sensitivity_beta}
\scriptsize
\setlength{\tabcolsep}{2.7pt}
\resizebox{\textwidth}{!}{%
\begin{tabular}{cc*{4}{cc}}
\toprule
$\beta$ & Cost regime & ADNI AUROC & ADNI Cost & WOMAC AUROC & WOMAC Cost & KLG AUROC & KLG Cost & ILIADD AUROC & ILIADD Cost \\
\midrule
0.5 & Low  & 0.7495 & 2.775  & 0.6217 & 8.966  & 0.8177 & 9.272  & 0.7941 & 19.425 \\
1   & Low  & 0.7487 & 2.760  & 0.6230 & 8.781  & 0.8164 & 9.390  & 0.7933 & 19.582 \\
2   & Low  & 0.7506 & 2.772  & 0.6228 & 9.055  & 0.8165 & 9.574  & 0.7965 & 19.868 \\
\midrule
0.5 & High & 0.7607 & 11.785 & 0.6250 & 18.074 & 0.8310 & 33.251 & 0.8448 & 90.956 \\
1   & High & 0.7520 & 11.640 & 0.6248 & 17.484 & 0.8374 & 37.567 & 0.8460 & 88.172 \\
2   & High & 0.7515 & 11.845 & 0.6231 & 17.391 & 0.8341 & 33.699 & 0.8459 & 88.761 \\
\bottomrule
\end{tabular}}
\end{table}

\begin{table}[!ht]
\centering
\caption{Test performance and average cost by embedding dimension $E$ and cost regime.}
\label{tab:sensitivity_embedding}
\scriptsize
\setlength{\tabcolsep}{2.7pt}
\resizebox{\textwidth}{!}{%
\begin{tabular}{cc*{4}{cc}}
\toprule
$E$ & Cost regime & ADNI AUROC & ADNI Cost & WOMAC AUROC & WOMAC Cost & KLG AUROC & KLG Cost & ILIADD AUROC & ILIADD Cost \\
\midrule
16  & Low  & 0.7487 & 2.763  & 0.6192 & 9.024  & 0.8195 & 9.766  & 0.7937 & 19.496 \\
32  & Low  & 0.7488 & 2.768  & 0.6235 & 8.965  & 0.8101 & 8.845  & 0.7956 & 19.441 \\
64  & Low  & 0.7487 & 2.760  & 0.6183 & 8.964  & 0.8178 & 9.582  & 0.7940 & 19.454 \\
128 & Low  & 0.7537 & 2.772  & 0.6234 & 8.902  & 0.8147 & 9.284  & 0.7980 & 19.328 \\
\midrule
16  & High & 0.7579 & 11.490 & 0.6232 & 17.288 & 0.8322 & 37.052 & 0.8460 & 88.529 \\
32  & High & 0.7537 & 11.922 & 0.6268 & 16.859 & 0.8353 & 39.903 & 0.8460 & 89.421 \\
64  & High & 0.7579 & 11.837 & 0.6236 & 17.378 & 0.8329 & 32.547 & 0.8481 & 90.211 \\
128 & High & 0.7543 & 12.390 & 0.6235 & 17.547 & 0.8351 & 35.930 & 0.8463 & 89.195 \\
\bottomrule
\end{tabular}}
\end{table}

\begin{table}[!ht]
\centering
\caption{Test performance and average cost by number of top plans $r$ and cost regime.}
\label{tab:sensitivity_r}
\scriptsize
\setlength{\tabcolsep}{2.7pt}
\resizebox{\textwidth}{!}{%
\begin{tabular}{cc*{4}{cc}}
\toprule
$r$ & Cost regime & ADNI AUROC & ADNI Cost & WOMAC AUROC & WOMAC Cost & KLG AUROC & KLG Cost & ILIADD AUROC & ILIADD Cost \\
\midrule
3  & Low  & 0.7493 & 2.808  & 0.6197 & 8.853  & 0.8134 & 9.190  & 0.7948 & 19.365 \\
5  & Low  & 0.7488 & 2.768  & 0.6254 & 9.104  & 0.8169 & 9.294  & 0.7918 & 19.248 \\
10 & Low  & 0.7468 & 2.788  & 0.6237 & 8.943  & 0.8189 & 9.719  & 0.7945 & 19.552 \\
\midrule
3  & High & 0.7545 & 11.700 & 0.6209 & 17.150 & 0.8359 & 36.646 & 0.8464 & 89.641 \\
5  & High & 0.7551 & 11.935 & 0.6265 & 17.277 & 0.8374 & 38.735 & 0.8466 & 87.570 \\
10 & High & 0.7545 & 11.787 & 0.6251 & 17.433 & 0.8354 & 35.513 & 0.8425 & 85.558 \\
\bottomrule
\end{tabular}}
\end{table}

\subsection{Candidate-Plan Set Size and Offline Construction}
\label{appendex:candidate_scalability}
We first evaluate how the candidate-set size $|\mathcal{P}|$ affects the performance--cost tradeoff. We varied the number of candidate plans $|\mathcal{P}| \in \{100, 500, 1000, 5000, 10000\}$ across datasets with different trajectory lengths ($L$) and feature dimensions ($M$). As shown in Table~\ref{tab:candidate_count_performance}, a small candidate set (e.g., $|\mathcal{P}|=100$) can yield less optimal plans, resulting in higher acquisition cost on datasets like KLG or ILIADD. Expanding the set to $|\mathcal{P}|=10000$ suggests diminishing returns, as additional computational expense does not translate into meaningful performance gains, supporting $|\mathcal{P}|=5000$ as a strong practical default.

\begin{figure}[!ht]
\centering
\includegraphics[width=0.75\textwidth]{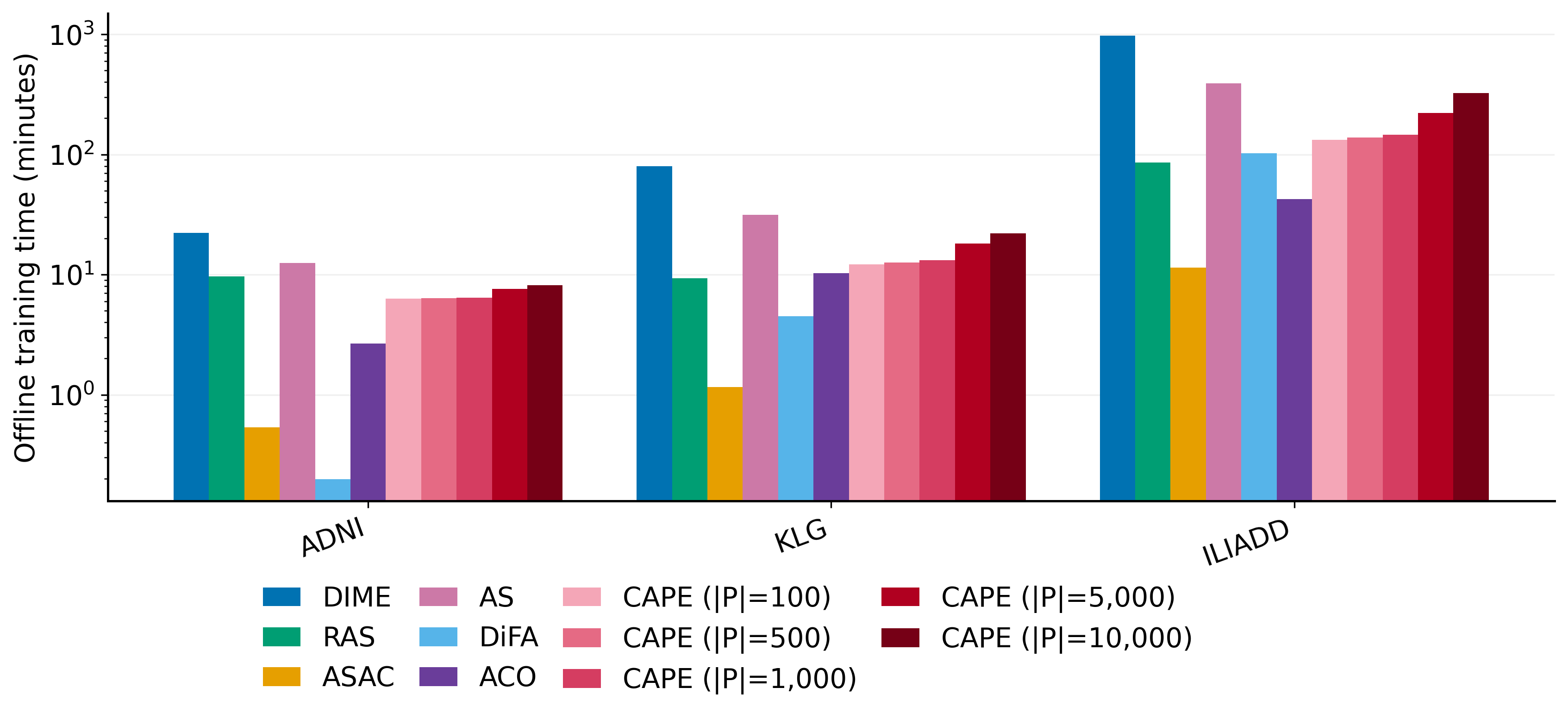}
\caption{Offline training time in minutes (lower is better).}
\label{fig:candidate_count_training}
\end{figure}

Figure~\ref{fig:candidate_count_training} shows the offline construction time across datasets with varying trajectory lengths ($L$) and feature dimensions ($M$). While runtime scales with $|\mathcal{P}|$, it remains substantially faster than baselines like DIME and AS across all datasets at default $|\mathcal{P}|=5000$. Importantly, this computation is strictly an offline policy-construction cost and does not impact \OurMethod's highly efficient nearest-centroid lookup during inference.

\begin{table}[!ht]
\centering
\caption{\OurMethod performance varying the number of candidate plans $|\mathcal{P}|$ across datasets.}
\label{tab:candidate_count_performance}
\small
\resizebox{\textwidth}{!}{%
\begin{tabular}{r*{4}{rr}}
\toprule
$|\mathcal{P}|$ & Synthetic Acc. & Synthetic Cost & ADNI AUROC & ADNI Cost & KLG AUROC & KLG Cost & ILIADD AUROC & ILIADD Cost \\
\midrule
100    & 0.8709 & 14.498 & 0.7485 & 11.772 & 0.8348 & 42.087 & 0.8367 & 105.824 \\
500    & 0.8673 & 14.113 & 0.7505 & 11.312 & 0.8349 & 35.249 & 0.8411 & 95.380 \\
1,000  & 0.8805 & 13.780 & 0.7498 & 11.677 & 0.8381 & 40.366 & 0.8401 & 94.587 \\
5,000  & 0.8845 & 13.647 & 0.7520 & 11.640 & 0.8374 & 37.567 & 0.8460 & 88.172 \\
10,000 & 0.8838 & 13.865 & 0.7503 & 11.738 & 0.8364 & 34.542 & 0.8481 & 86.389 \\
\bottomrule
\end{tabular}}
\end{table}

\subsection{Offline Training Cost and Scalability}
\label{appendex:offline_scalability}
We evaluate the offline scalability of \OurMethod using a controlled setup on the Intensive Longitudinal Investigation of Alternative Diagnostic Dimensions (ILIADD) dataset. In the main experiments, we used a trajectory length of $L=10$ with $M=34$ features to predict negative affect. We evaluate scalability along two axes:
\begin{itemize}[leftmargin=*]
    \item \textbf{Scaling Time ($L$):} We fixed $M$ and increased the time length to $L \in \{20,30,50\}$, scaling up to 1,700 (feature, time) pairs.
    \item \textbf{Scaling Features ($M$):} We fixed $L=20$ and increased the feature dimensions to $M \in \{68,102\}$, scaling up to 2,040 (feature, time) pairs. To do this, we retained the original 34 features and appended standard Gaussian noise features.
\end{itemize}

\begin{figure}[!ht]
\centering
\includegraphics[width=0.63\textwidth]{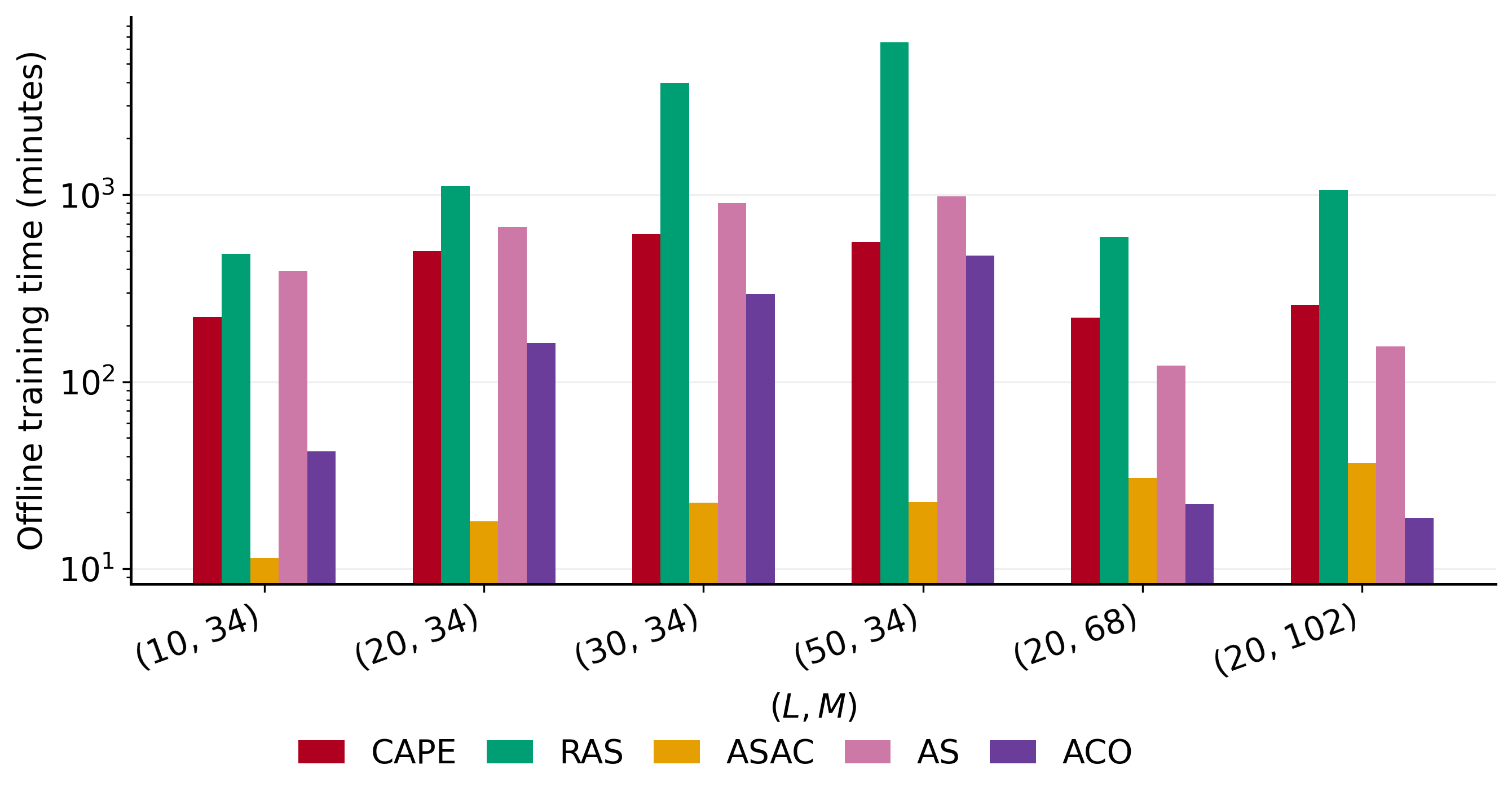}
\caption{ILIADD offline training time across methods by $(L,M)$, in minutes. Note that $L$ is the number of timesteps and $M$ is the number of features.}
\label{fig:iliadd_scaling_training}
\end{figure}

\begin{figure}[!ht]
\centering
\includegraphics[width=0.63\textwidth]{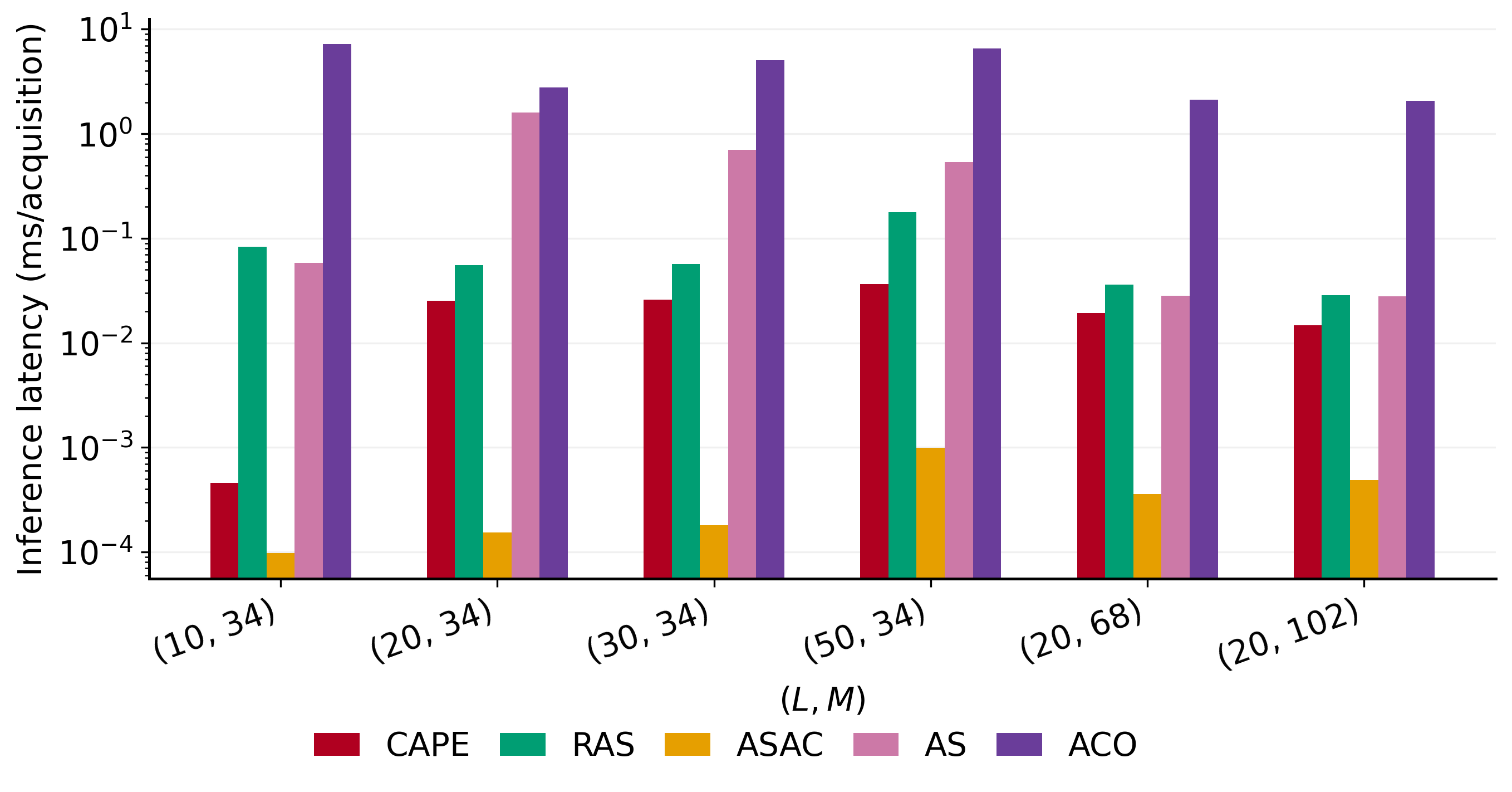}
\caption{ILIADD inference latency across methods by $(L,M)$, in milliseconds per acquisition. Note that $L$ is the number of timesteps and $M$ is the number of features.}
\label{fig:iliadd_scaling_inference}
\end{figure}

As shown in Table~\ref{tab:iliadd_scaling_performance} and Figures~\ref{fig:iliadd_scaling_training}--\ref{fig:iliadd_scaling_inference}, \OurMethod scales to much larger horizons and dimensionalities. Across every $(L,M)$ combination, \OurMethod achieves the highest AUROC and the lowest acquisition cost. Furthermore, \OurMethod's offline construction time grows predictably and is comparable to other baselines (avoiding the substantial runtime growth that limits baselines like RAS at high $L$), all while maintaining sub-millisecond inference latency per acquisition.

\begin{table}[!ht]
\centering
\caption{ILIADD test AUROC and average acquisition cost across methods by $(L,M)$.}
\label{tab:iliadd_scaling_performance}
\scriptsize
\setlength{\tabcolsep}{3pt}
\resizebox{\textwidth}{!}{%
\begin{tabular}{rr*{5}{rr}}
\toprule
$L$ & $M$ & RAS AUROC & RAS Cost & ASAC AUROC & ASAC Cost & AS AUROC & AS Cost & ACO AUROC & ACO Cost & \OurMethod AUROC & \OurMethod Cost \\
\midrule
10 & 34  & 0.5780 & 41.097  & 0.6328 & 28.600  & 0.7415 & 204.497 & 0.7789 & 53.147  & \textbf{0.7862} & \textbf{16.672} \\
20 & 34  & 0.7507 & 283.176 & 0.6309 & 109.795 & 0.7308 & 109.639 & 0.7517 & 111.830 & \textbf{0.8294} & \textbf{42.338} \\
30 & 34  & 0.7579 & 441.695 & 0.6710 & 110.000 & 0.7224 & 109.756 & 0.7468 & 110.168 & \textbf{0.8376} & \textbf{56.791} \\
50 & 34  & 0.7492 & 1508.057& 0.5641 & 40.645  & 0.6969 & 109.785 & 0.6872 & 33.475  & \textbf{0.7621} & \textbf{25.844} \\
20 & 68  & 0.7146 & 442.117 & 0.6592 & 109.992 & 0.6567 & 187.338 & 0.6539 & 101.002 & \textbf{0.7764} & \textbf{49.605} \\
20 & 102 & 0.6555 & 725.170 & 0.6372 & 109.930 & 0.6008 & 267.578 & 0.6618 & 54.408  & \textbf{0.6912} & \textbf{52.451} \\
\bottomrule
\end{tabular}}
\end{table}

\newpage
\subsection{Average Precision Evaluation}
\label{appendex:prauc}
\begin{figure}[!ht]
\centering
\includegraphics[width=\textwidth]{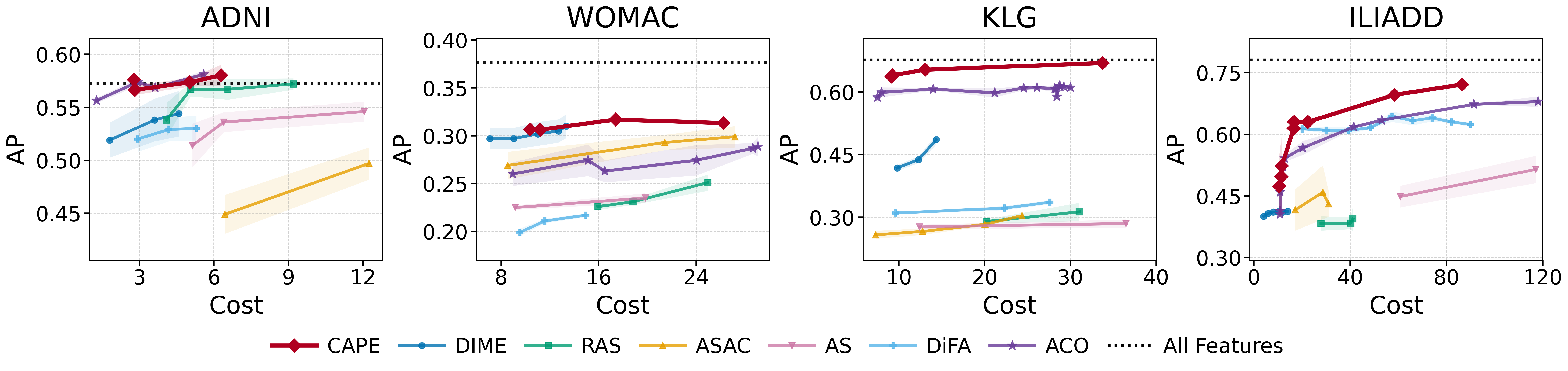}
\caption{Average precision (AP) versus average acquisition cost (budget) of \OurMethod and the baselines. The dotted horizontal line shows performance using all features.}
\label{fig:prauc}
\end{figure}

We evaluate average precision (AP), detailed in Figure~\ref{fig:prauc}. The results demonstrate that \OurMethod consistently achieves higher AP while incurring lower acquisition costs compared to the baselines. These results confirm that the strong Pareto-frontier performance observed in AUROC translates directly to AP.

\newpage
\subsection{Additional Acquisition-Embedding Visualizations}

\begin{figure}[!ht]
\centering
    \begin{subfigure}[t]{0.48\linewidth}
        \centering
        \includegraphics[width=\linewidth]{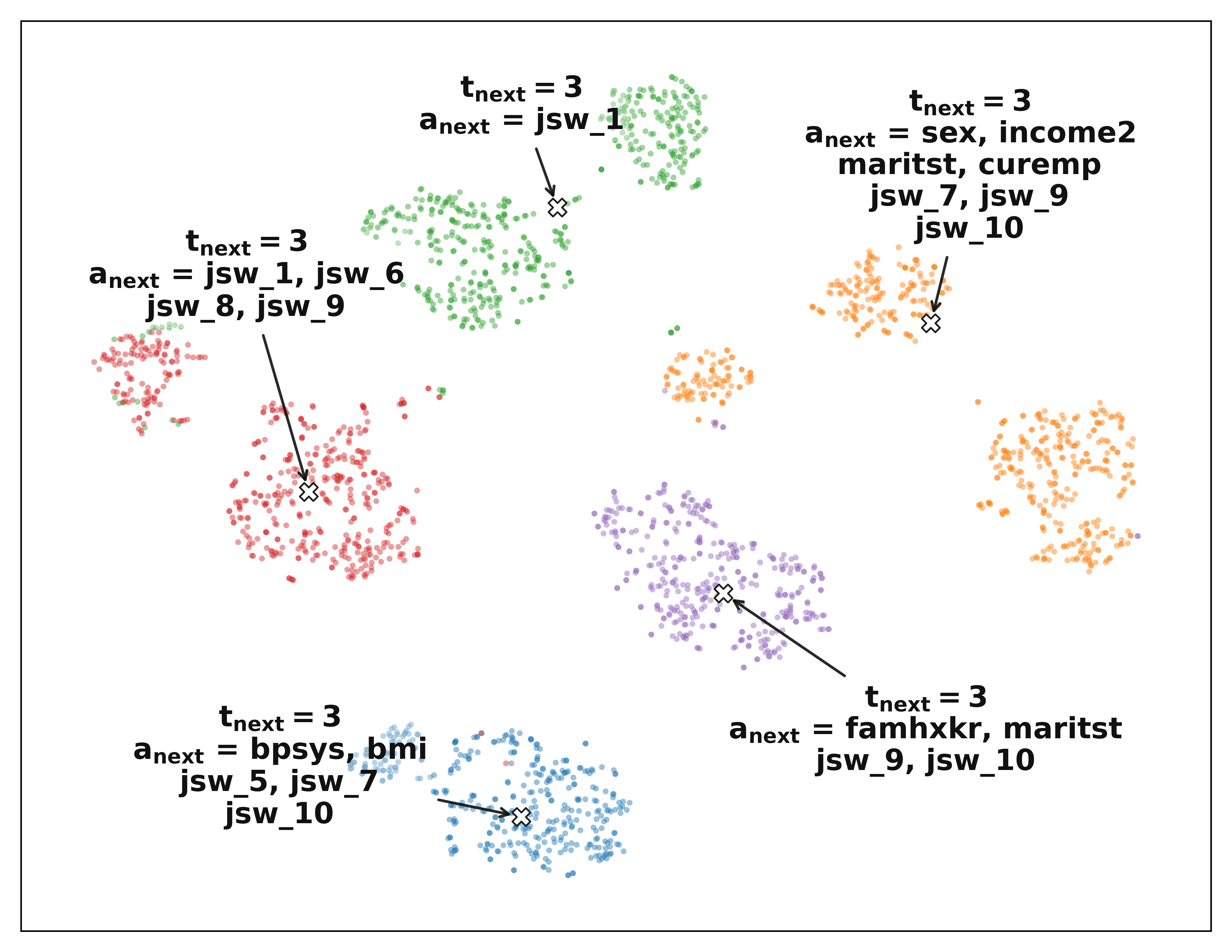}
        \caption{Acquisition regions at time $t=2$.}
        \label{fig:appx_tsne_t3}
    \end{subfigure}
    \hfill
    \begin{subfigure}[t]{0.48\linewidth}
        \centering
        \includegraphics[width=\linewidth]{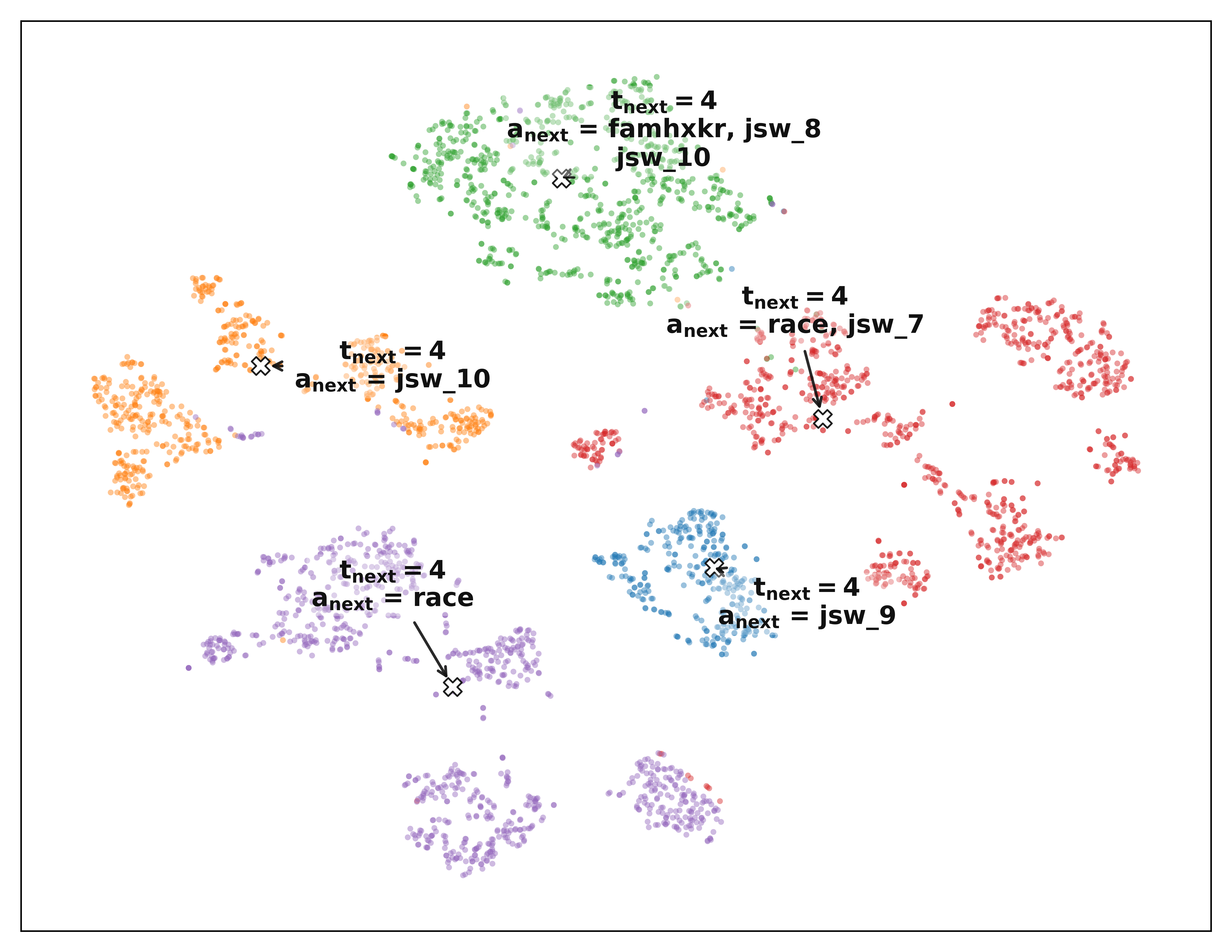}
        \caption{Acquisition regions at time $t=3$.}
        \label{fig:appx_tsne_t4}
    \end{subfigure}
    
    \vspace{0.5cm}
    
    \begin{subfigure}[t]{0.48\linewidth}
        \centering
        \includegraphics[width=\linewidth]{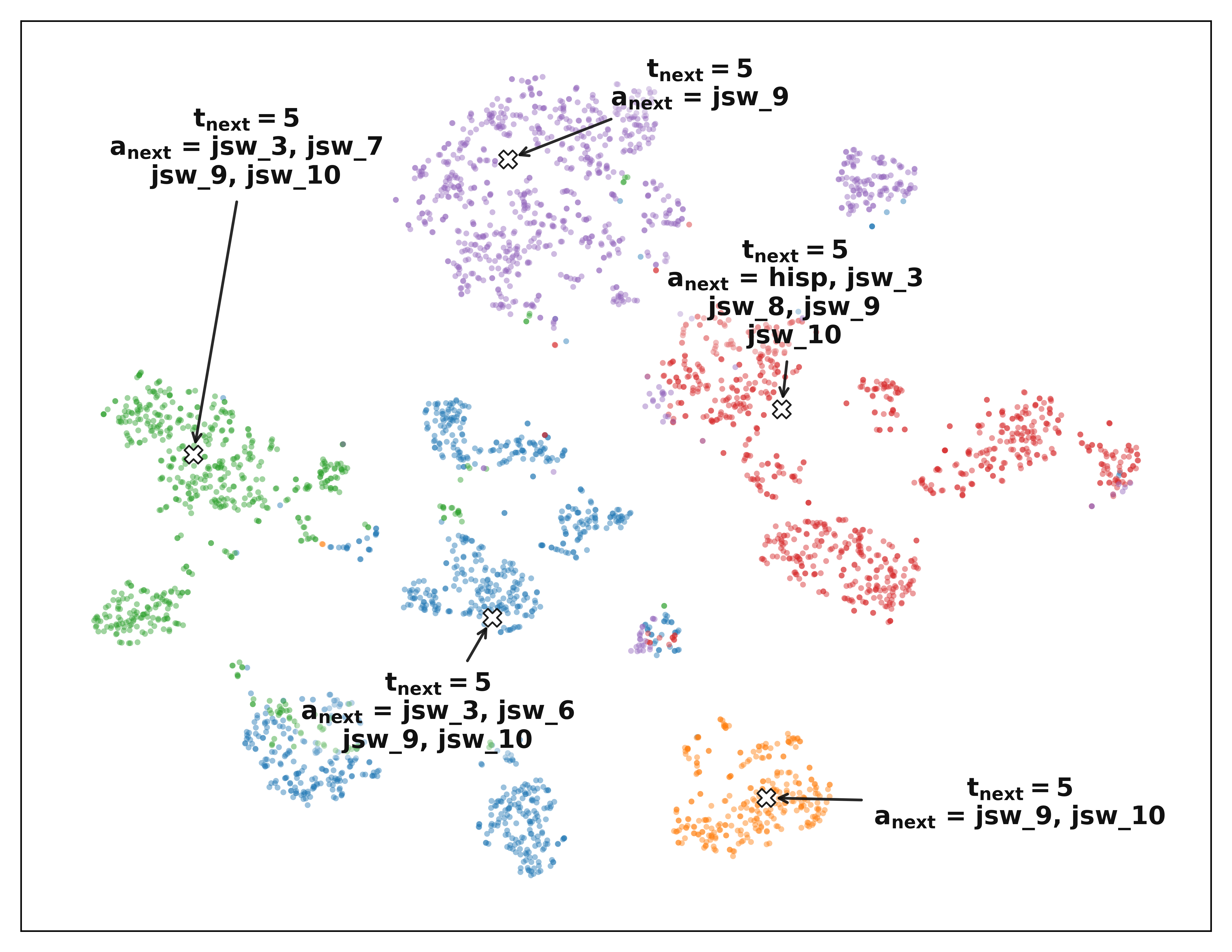}
        \caption{Acquisition regions at time $t=4$.}
        \label{fig:appx_tsne_t5}
    \end{subfigure}
    \hfill
    \begin{subfigure}[t]{0.48\linewidth}
        \centering
        \includegraphics[width=\linewidth]{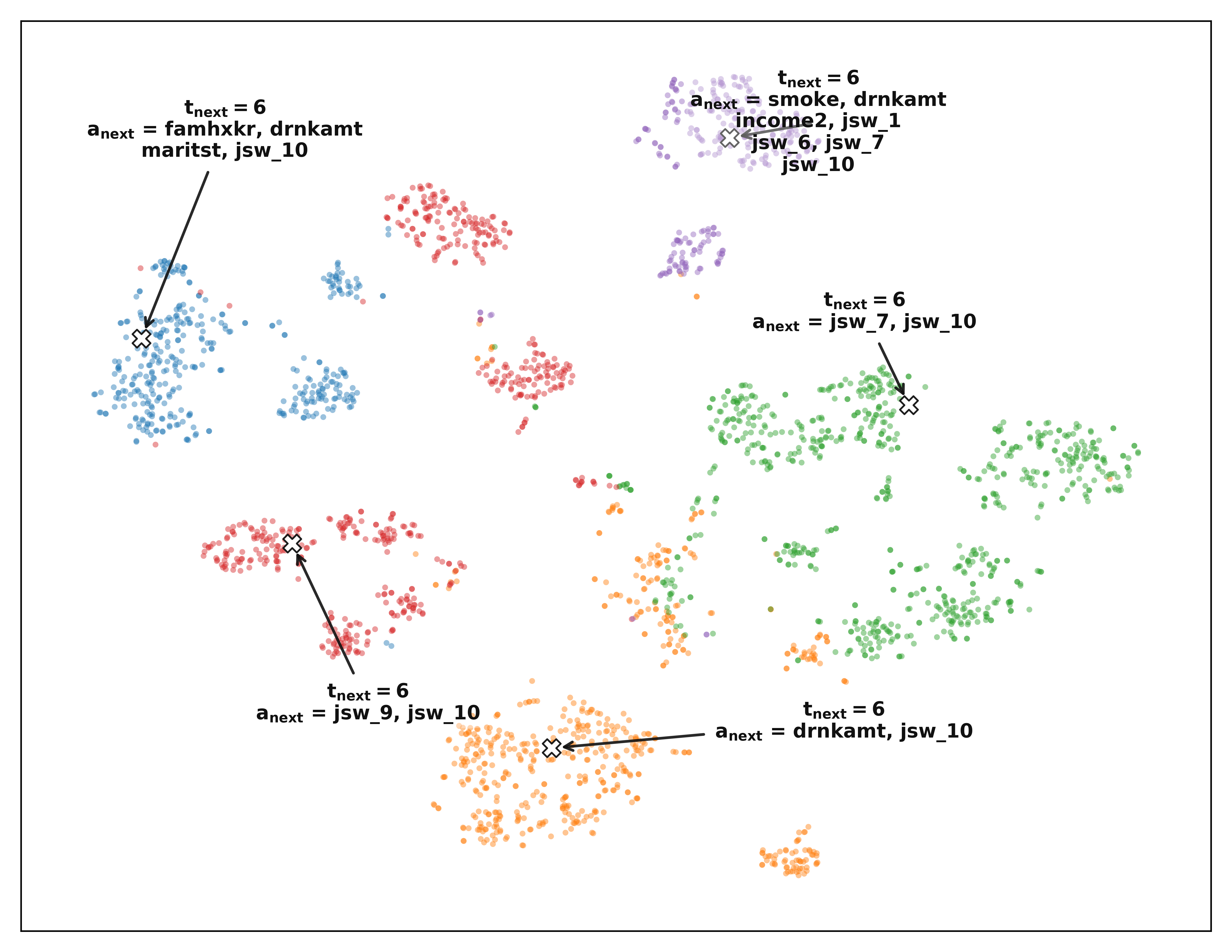}
        \caption{Acquisition regions at time $t=5$.}
        \label{fig:appx_tsne_t6}
    \end{subfigure}
    \caption{
    Extended visualizations of the acquisition-aware embeddings on the KLG task at later rollout stages. Building upon Figure \ref{fig:embedding_diagnostic}, the t-SNE projections at rollout times $t \in \{2, 3, 4, 5\}$ demonstrate how the learned representation maintains distinct local regions for shared cached acquisition decisions as the trajectory progresses.
    }
    \label{fig:appx_embedding_regions_extended}
\end{figure}